\documentclass[10pt,twocolumn,letterpaper]{article}

\usepackage[pagenumbers]{cvpr} 

\usepackage{graphicx}
\usepackage{amsmath}
\usepackage{amssymb}
\usepackage{booktabs}

\usepackage[pagebackref,breaklinks,colorlinks]{hyperref}

\usepackage[capitalize]{cleveref}
\crefname{section}{Sec.}{Secs.}
\Crefname{section}{Section}{Sections}
\Crefname{table}{Table}{Tables}
\crefname{table}{Tab.}{Tabs.}

\def\cvprPaperID{9244} 
\def\confName{CVPR}
\def\confYear{2022}

\usepackage{multirow}
\usepackage{amsmath}

\usepackage{microtype}
\usepackage{graphicx}
\usepackage{caption}
\usepackage{appendix}
\usepackage{subcaption}
\usepackage{algorithm}
\usepackage{amsfonts}

\usepackage{amsmath,amsthm,amssymb}
\usepackage[noend]{algpseudocode}

\newtheorem{proposition}{Proposition}

\theoremstyle{definition}

\DeclareMathAlphabet\mathbfcal{OMS}{cmsy}{b}{n}

\usepackage[noend]{algpseudocode}
\newcommand{\noref}[1]{\ref{#1}}

\DeclareGraphicsExtensions{.png,.jpg,.pdf}

\begin{document}
\title{Implicit Feature Decoupling with Depthwise Quantization}

\author{Iordanis Fostiropoulos\\
University of Southern California\\
Los Angeles, CA\\
{\tt\small fostirop@usc.edu}
\and
Barry Boehm\\
University of Southern California\\
Los Angeles, CA\\
{\tt\small boehm@usc.edu}
}
\maketitle

\begin{abstract}
Quantization has been applied to multiple domains in Deep Neural Networks (DNNs). We propose Depthwise Quantization (DQ) where \textit{quantization} is applied to a decomposed sub-tensor along the \textit{feature axis} of weak statistical dependence. The feature decomposition leads to an exponential increase in \textit{representation capacity} with a linear increase in memory and parameter cost. In addition, DQ can be directly applied to existing encoder-decoder frameworks without modification of the DNN architecture. We use DQ in the context of Hierarchical Auto-Encoders
and train end-to-end on an image feature representation. We provide an analysis of the cross-correlation between spatial and channel features and propose a decomposition of the image feature representation along the channel axis. The improved performance of the depthwise operator is due to the increased representation capacity from implicit feature decoupling.  We evaluate DQ on the likelihood estimation task, where it outperforms the previous state-of-the-art on CIFAR-10, ImageNet-32 and ImageNet-64. We progressively train with increasing image size a single hierarchical model that uses 69\% fewer parameters and has faster convergence than the previous work. 
\end{abstract}

\section{Introduction}

 \begin{figure}[ht]
\centering
\begin{minipage}[h]{0.90\linewidth}

         \includegraphics[width=1\textwidth]{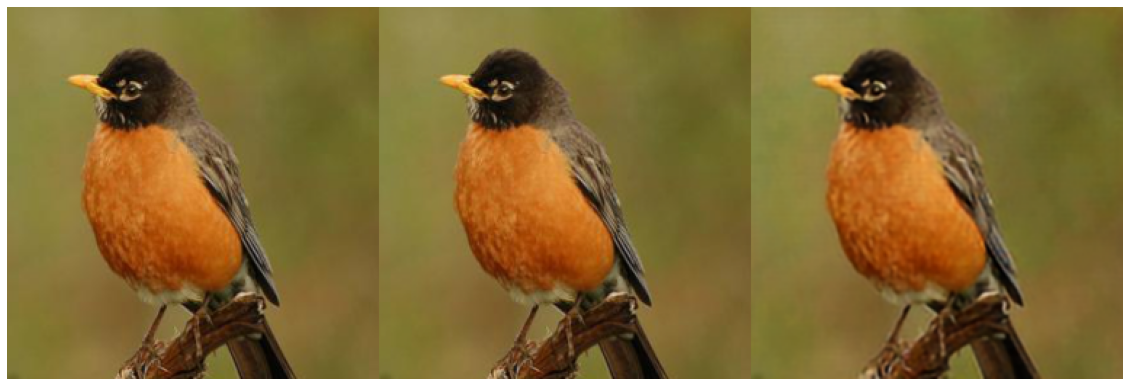}

\end{minipage}

\begin{minipage}[h]{0.90\linewidth}
         \includegraphics[width=1\textwidth]{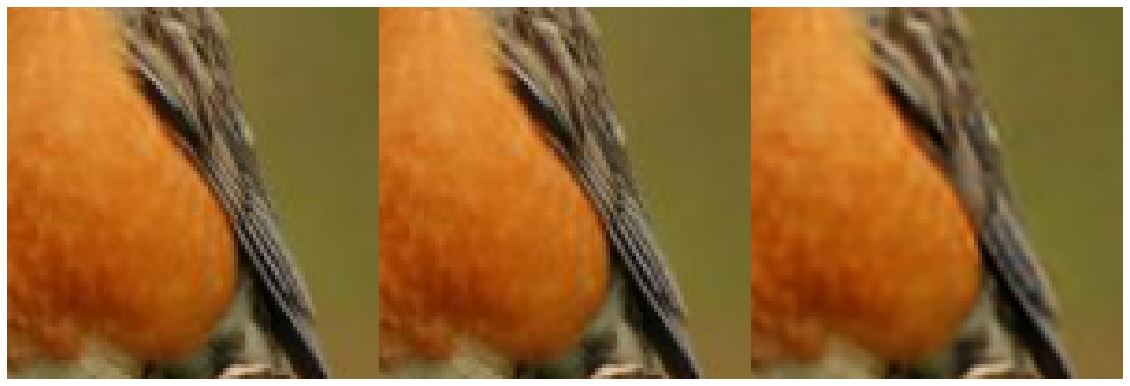}
\end{minipage}

\begin{minipage}[h]{0.90\linewidth}

         \includegraphics[width=1\textwidth]{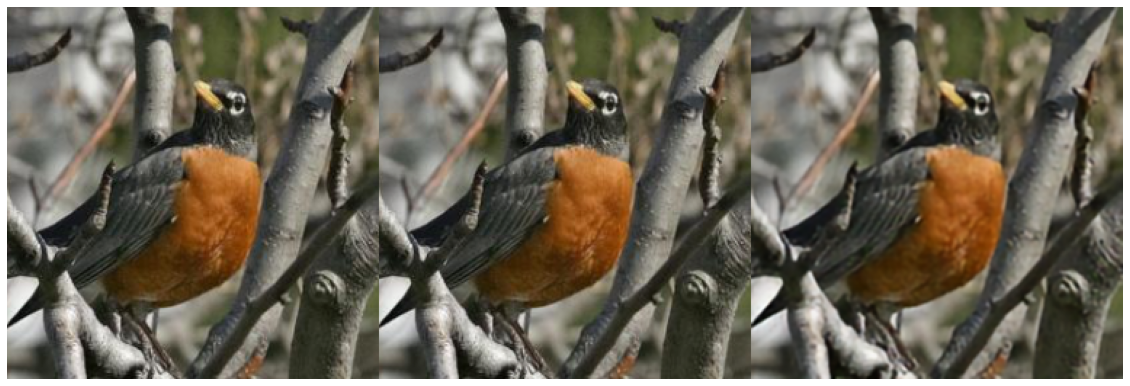}

\end{minipage}
\begin{minipage}[h]{0.90\linewidth}
         \includegraphics[width=1\textwidth]{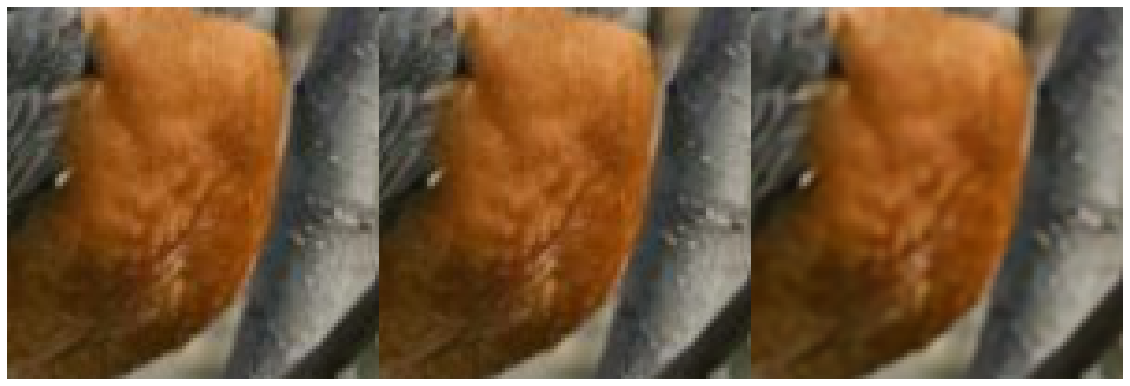}
\end{minipage}

\centering         
\caption{The original images (left) are reconstructed by DQ (middle) and VQ (right) with identical models and training setup. The perceptual quality of DQ outperform VQ.}

\label{fig:comparison}
 \end{figure}

Quantization is an effective lossy compression process that maps a continuous signal to a set of discrete values, also called \textit{codes}. Quantization is extended to vector feature spaces with learning paradigms such as \textit {Vector Quantization} (VQ) and with a training objective identical to k-means. \textit{Product Quantization} (PQ) decomposes the feature vector and assumes a weak statistical dependence between feature sub-vectors.  \textit{Additive Quantization} (AQ) decomposes the feature vector into a sum of quantized vectors as opposed to the concatenated output in PQ.

\begin{figure*}[ht]
\centering
\begin{subfigure}{.95\textwidth}
\includegraphics[width=1\textwidth]{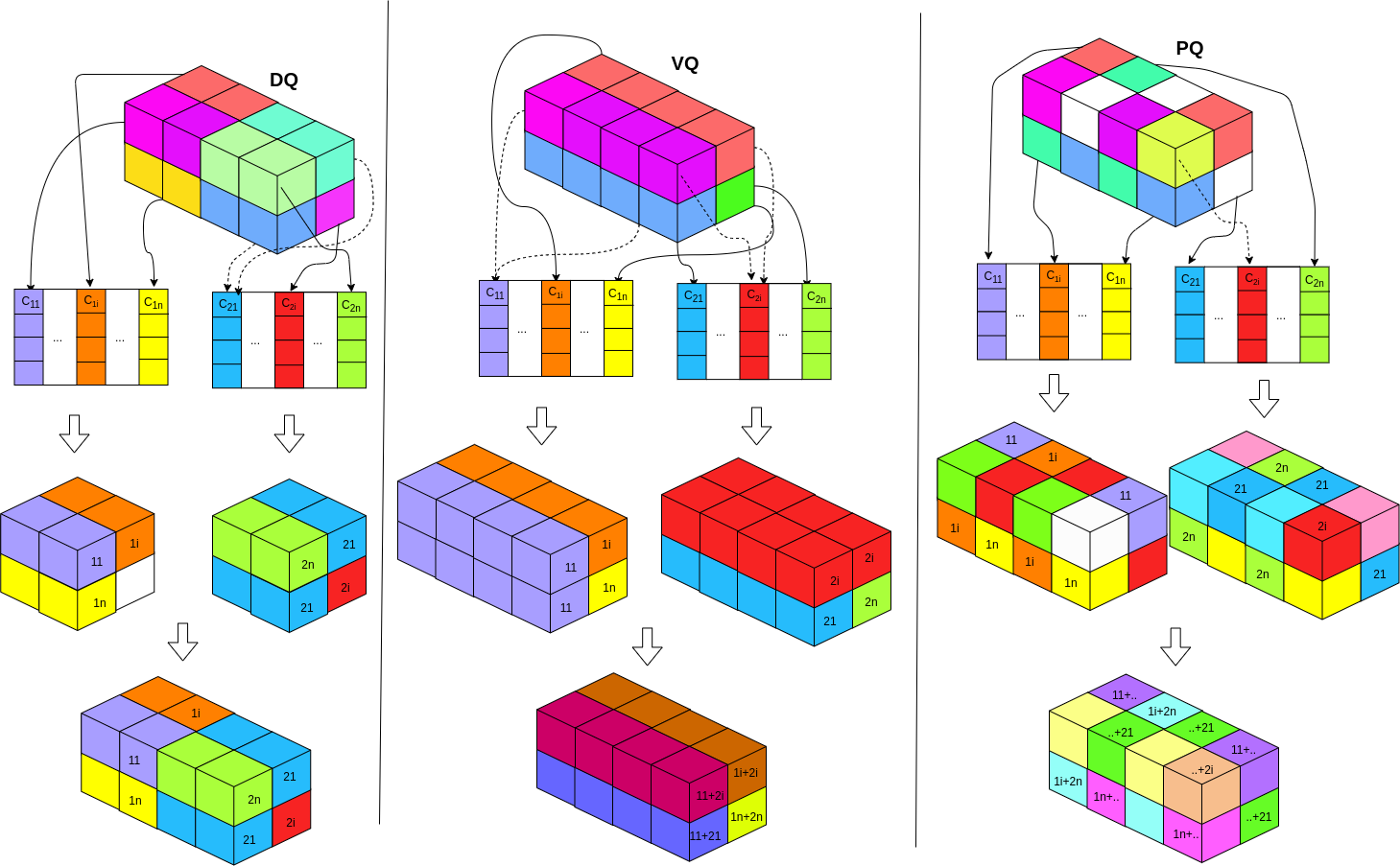}
\end{subfigure}
\centering
\caption{
DQ (left) apply $C_1$ on the first slice of the sub-tensor and for all sub-vectors and concatenate the quantized vectors. VQ \cite{vqvae} (middle) and PQ (right) quantize the same vector with different codebook and combine the two sub-vectors by addition or concatenation.}
     \label{fig:architecture}
 \end{figure*}
 
Quantization is used in conjunction with Deep Neural Networks (DNNs) for tasks such as classification\cite{prototype_cnn}, incremental learning\cite{prototype_cnn}, zero-shot learning \cite{zero_shot_prototype}, generation \cite{vqvae}, compression \cite{image_compression} and data retrieval \cite{cao_image_retrieval}. The discrete quantized feature representations can be used post-hoc \cite{vqvae,vqvae2,jukebox} or as a learning objective (i.e. classification) \cite{zero_shot_prototype,prototype_cnn}.  Our work is motivated by the increasing number of quantization applications to high dimensional feature tensors. We view the quantizer as a density estimator and evaluate it on the task of likelihood estimation for the visual domain. 

Likelihood estimation models seek to minimize the \textit{divergence} between the data distribution and the model prior. Explicit likelihood estimation models, including Vector Quantization (VQ), Variational Auto-encoders (VAEs), and Auto-Regressive (AR) models directly minimize a divergence. In this work, we focus on explicit likelihood estimation.  

AR models do well in likelihood estimation and are applied in multiple domains such as language, vision, and audio. AR models have a recursive dependency on the input during training and inference. Therefore, AR models are computationally inefficient for domains with long sequences, such as pixels of an image. Even with caching \cite{cache_wavenet} during sampling, AR models are still less efficient than VAEs. 

The priors of VAEs provide a compressed feature representation that can be used as a surrogate training objective for the downstream task. In contrast to the discrete prior, a continuous prior can lead to \textit{posterior collapse}. The representation is ignored by the downstream task model because it is either too noisy or uninformative. This effect is amplified when the data is discrete, as in  the language domain\cite{z_forcing}.


To that end, we propose the Depthwise Quantization (DQ) method that quantizes each decomposed feature sub-tensor with a different quantizer. We use rate-distortion theory to interpret a quantizer as an encoding function with limited capacity. We provide a theoretical upper bound on the capacity in relation to the quantization cost when DQ is applied on a decoupled feature tensor, as opposed to a coupled feature tensor. We evaluate the performance of DQ on the feature space of ImageNet for an image classification backbone. Lastly, we apply DQ to a hierarchical Auto-Encoder with DQ as a bottleneck for different hierarchies and train it end-to-end. DQ outperforms explicit models in likelihood estimation. In detail:

\begin{itemize}

\item We propose Depthwise Quantization (DQ) and decompose a feature tensor along the axis of weak statistical dependence. 

\item We provide a theoretical analysis on the improved quantization performance and experimentally corroborate our theoretical results. 


\item We introduce an improved hierarchical AutoEncoder model Depth-Quantized Auto-Encoder where DQ is applied to the feature representation at different hierarchies. 



\item We extend the parametric Mutual Information (MI) quantization estimators for DNNs when the prior is \textit{learned}. We experimentally verify that the learned prior is \textit{implicitly decoupled}.

\end{itemize}

Our approach can be applied to previous works that use quantization. We demonstrate with our experiments that DQ performs significantly better when the assumption on cross-correlation is strong in both post-hoc analysis and end-to-end training settings. When trained end-to-end, DQ reduces the cross-correlation among the decomposed feature tensors (``implicitly decouple'') and improves reconstruction loss and likelihood estimation. Our code is publicly available\footnote{\url{https://github.com/fostiropoulos/Depthwise-Quantization}}.

\section{Related Work}

Our work is closely related to previous studies on feature decomposition and quantization optimization in visual tasks using DNNs.

Feature decomposition approaches include \textit{Separable Convolutions} (SP) \cite{seperable_convnets}, which factorizes a convolutional kernel to the \textit{spatial} dimensions.
SP reduces the number of computations required to calculate the filter output. Inception \cite{inception}, another approach to feature decomposition, factorizes a feature representation implicitly with a ``Network-in-Network'' (NiN) \cite{network_in_network} branch of convolutions. Thus, Inception learns spatial cross-correlations and feature cross-correlations independently. 
\textit{Depthwise} Separable Convolution (DSC) \cite{xception,inception} is a method of feature decomposition that uses a single ``spatial'' convolution followed by multiple vanilla convolutions on a decomposed ``segment''. Xception\cite{xception} is based on the ``Inception Hypothesis'' for a decoupled space where DSC is applied to an \textit{e\textbf{X}treme}.
Our work is based on a hypothesis similar to that of DSC and considers modeling cross-channel correlations and spatial correlations independently. 

There are analyses on the \textit{decoupling} of the feature space \textit{implicitly} in the context of DSC as well as on NiN architectures. Blueprint Separable Convolutions (BSConv) \cite{bsconv} have been proposed as an alternative to DSC based on the observation of intra-kernel correlations. They propose a \textit{pointwise} (1x1) convolution followed by a depthwise convolution. In contrast, DSC enforces cross-kernel correlations \textit{implicitly}. Analysis on the variance of a convolution kernel shows that a DNN can perform better when cross-kernel redundancies decrease. 

Other works \textit{explicitly} factorize a convolutional filter. There are methods that use a low-rank approximation \cite{low_rank_aprox_conv} or closed-form decomposition \cite{network_decoupling} on pre-trained networks to speed up the computation process. Previous works on speeding up networks \cite{quantization_kernel} have used product quantization to quantize convolutional filters and take advantage of the redundancies. Previous analysis of the redundancy and cross-correlation of the feature space in DNNs is complementary to our work.

Improvements in quantization learning approaches include Optimized Product Quantization \cite{optimized_pq} that decomposes the feature vector in a parametric manner. In addition, Additive Quantization \cite{additive_quantization} improves on the computational efficiency of PQ for high dimensional vector search by decomposing the vectors into a sum instead of a concatenation of their sub-vectors. In contrast, our work can be applied to feature tensors and the quantizer is trained end-to-end with a DNN.

Kobayashi \etal \cite{decomposing_medical_images} train a quantizer end-to-end with a DNN. They use multiple codebooks and train each codebook independently for a different supervised task. However, as opposed to our method, the codebooks are decoupled in a supervised manner. Moreover, the quantized representations are used by different networks for different downstream tasks as opposed to interacting for a single downstream task. Lastly, vector decomposition is applied to feature vectors as opposed to feature sub-tensors as in our work. 

PQ-VAE \cite{pqvae} also decomposes the latent representations to sub-vectors and uses different quantizers for each sub-vector. Kaiser \etal \cite{fast_decoding} introduces ``sliced quantization'' that is identical to PQ-VAE but uses the discrete representation post-hoc with a latent variable model. By contrast, DQ decomposes the feature space to sub-tensors as opposed to sub-vectors and thus improves the reconstruction loss by implicitly increasing the statistical dependence within the sub-tensor. 

The works most similar to ours are those of Razavi \etal \cite{vqvae2} and Dhariwal \etal \cite{jukebox}. VQ-VAE-2 \cite{vqvae2} applies quantization to the feature representation of multiple hierarchies on a VAE. Similar to our method, they train VQ end-to-end with a VAE. However, we apply DQ as opposed to VQ for the quantization method. VQ-VAE-2 can suffer from an uninformative top prior. Subsequent models such as ``Jukebox'' \cite{jukebox} mitigate the issue by modeling each hierarchy with an independent encoder-decoder architecture. We also avoid the issue of an uninformative top prior but do not model each hierarchy with a different model. Instead, we introduce a model architecture DQ-AE. 

\section{Background}
\label{sec:background}

 \begin{figure*}[ht]
\centering

\begin{subfigure}{0.20\textwidth}
\begin{minipage}[h]{\textwidth}

         \includegraphics[width=\textwidth]{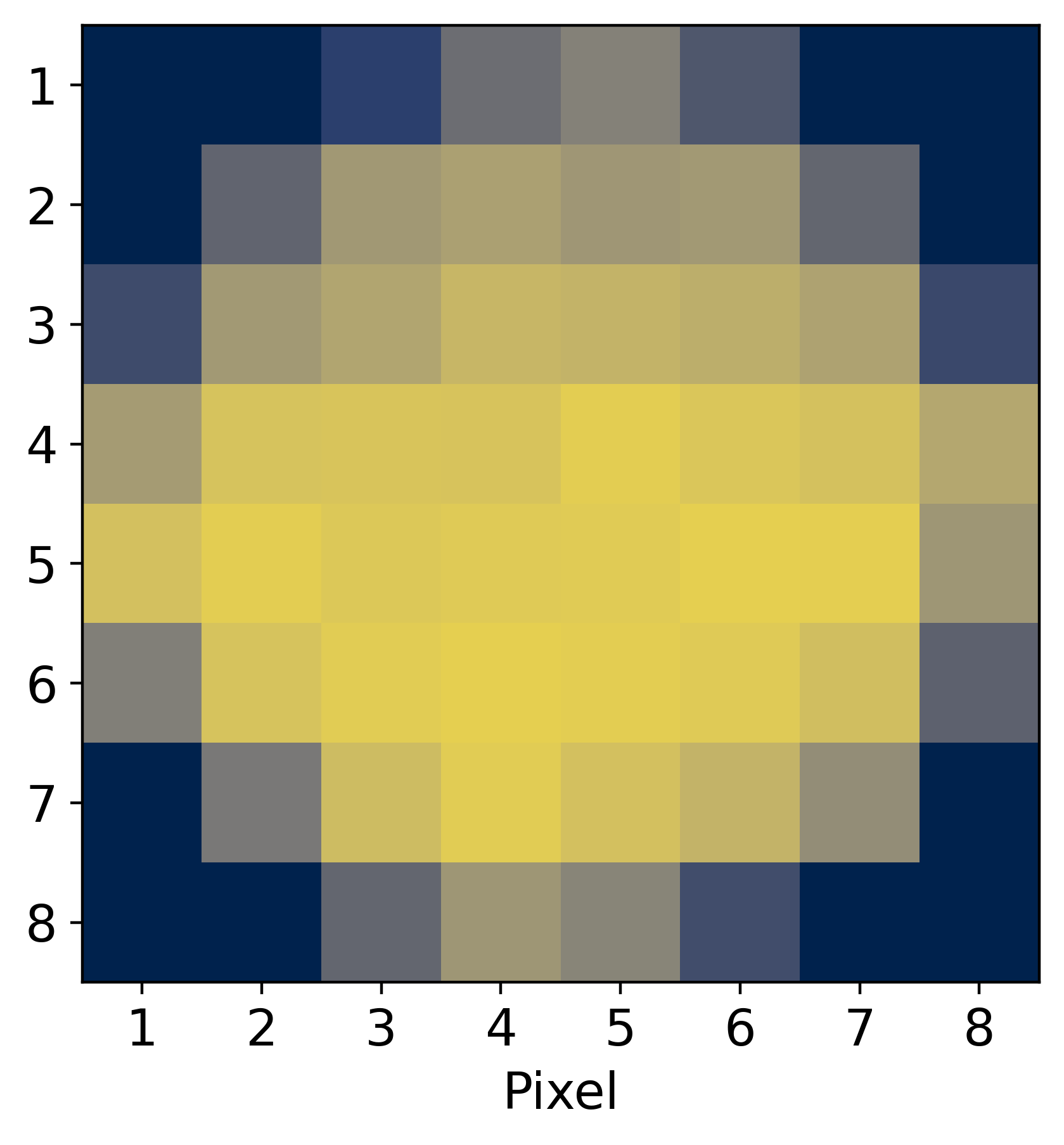}
         \label{fig:dvq_entropy_feature_map}
         
\end{minipage}
\end{subfigure}
\begin{subfigure}{0.26\textwidth}
\begin{minipage}[h]{\textwidth}
         \includegraphics[width=\textwidth]{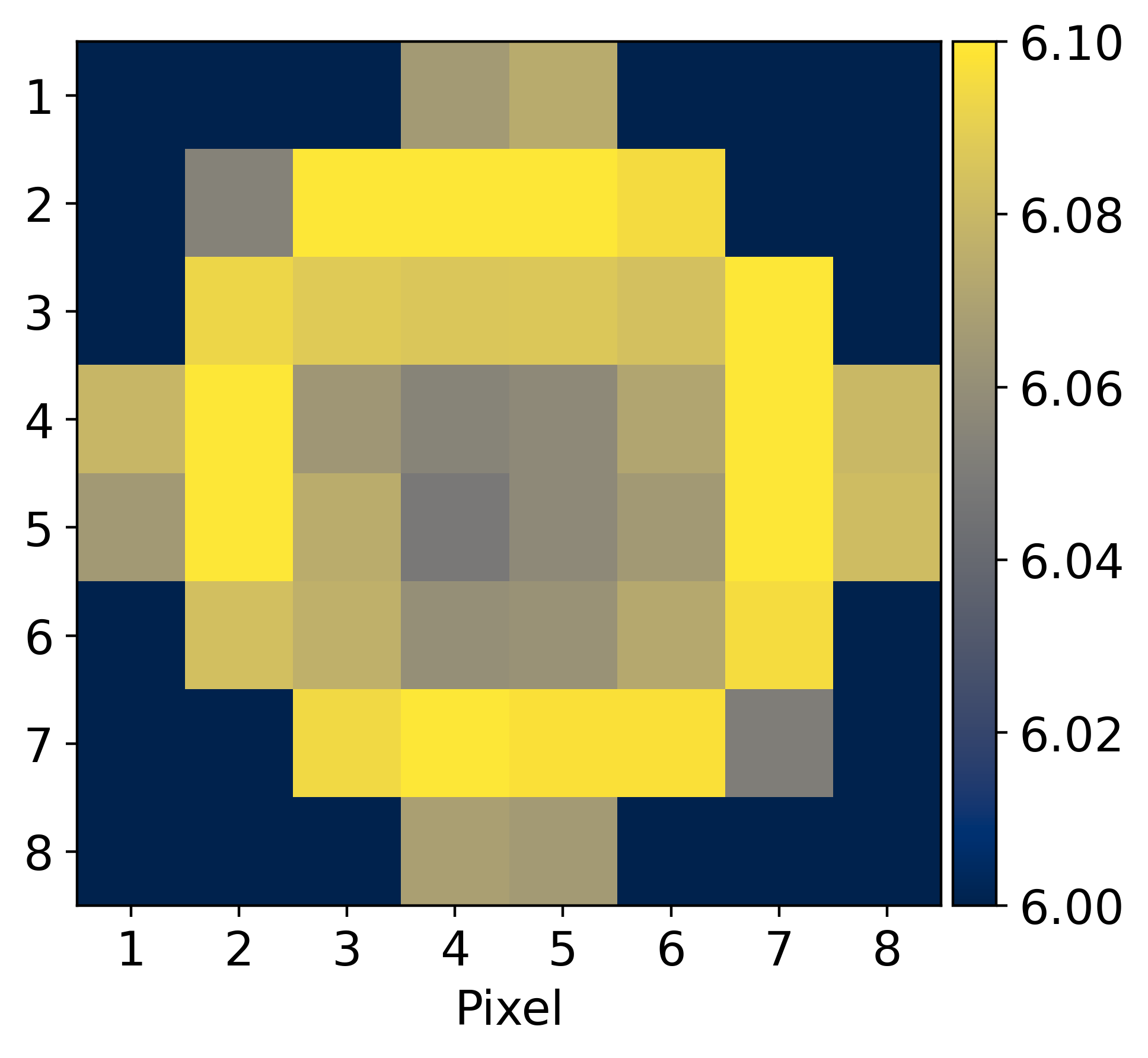}
         \label{fig:vq_entropy_feature_map}
         
\end{minipage}
\end{subfigure}
\begin{subfigure}{0.20\textwidth}
\begin{minipage}[h]{\textwidth}

         \includegraphics[width=\textwidth]{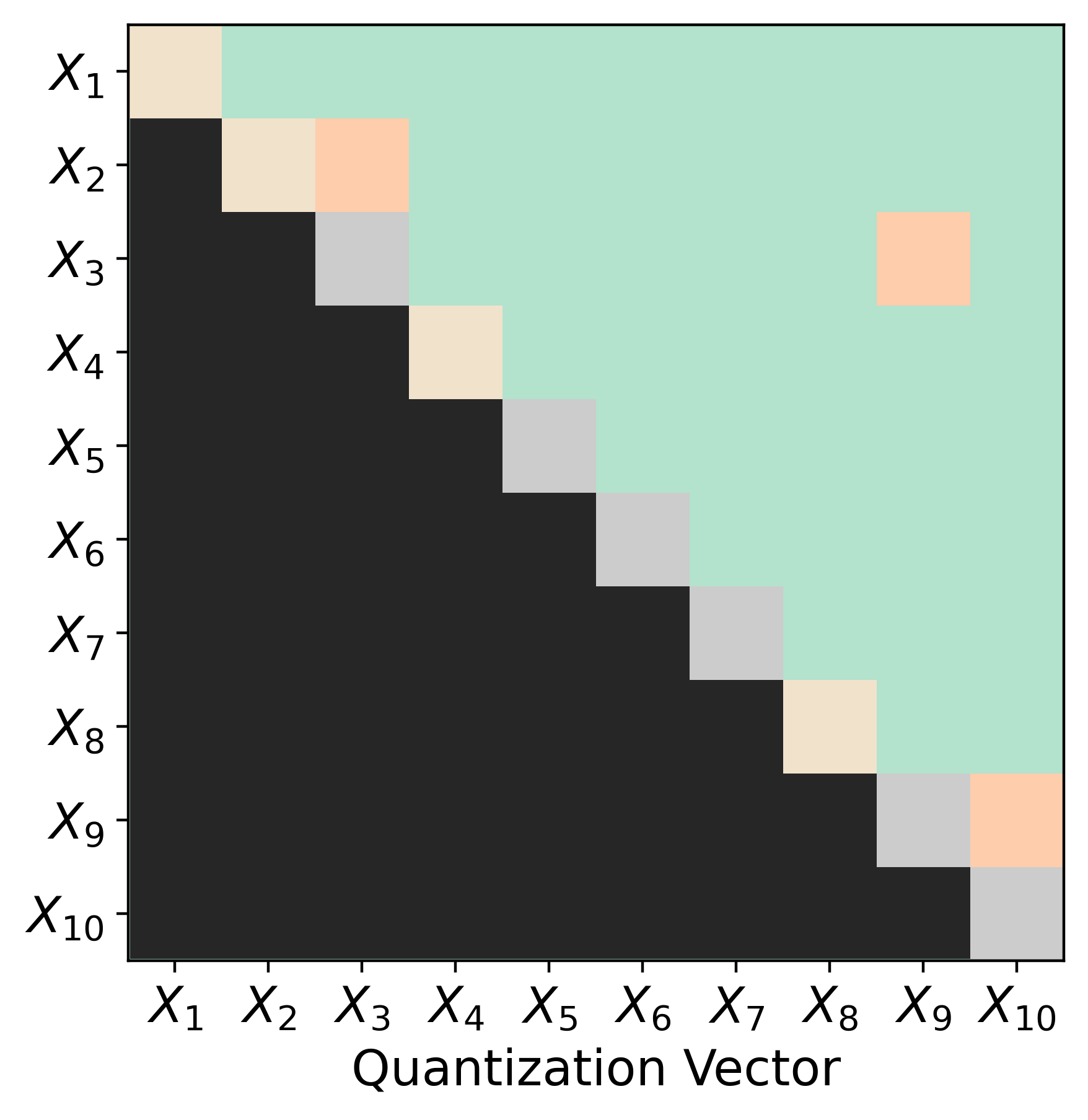}
         \label{fig:dvq_mi_map}

\end{minipage}
\end{subfigure}
\begin{subfigure}{0.25\textwidth}
\begin{minipage}[h]{\textwidth}

         \includegraphics[width=\textwidth]{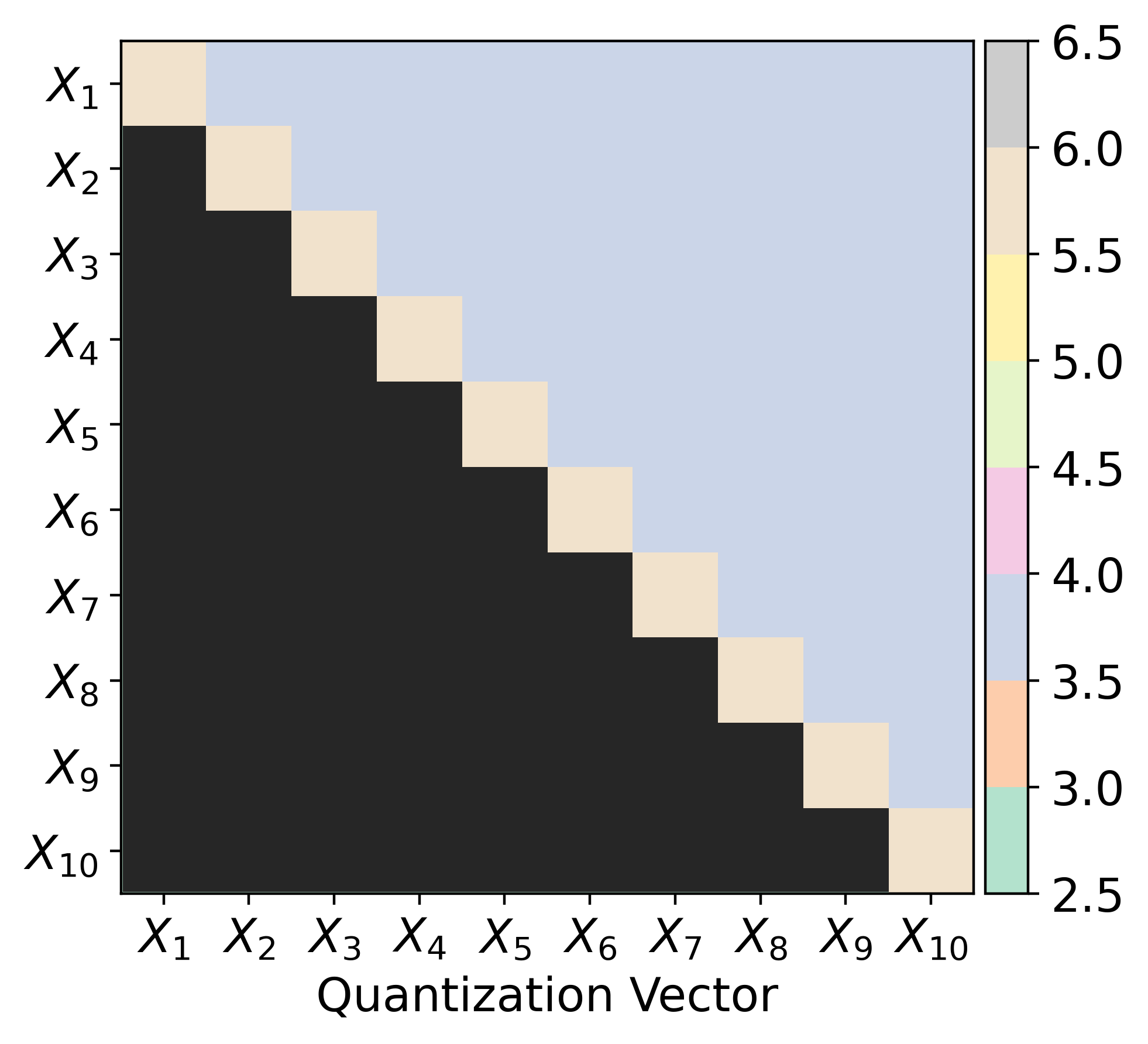}
         \label{fig:vq_mi_map}
\end{minipage}
\end{subfigure}
\centering
 \caption{Effects of implicit marginalization on the learned quantized features. The mean entropy (``informativeness'') of the quantization vectors for each pixel for DQ (first) is higher when compared to VQ (second); higher is better. The mean MI (``redundancy'') scores between each quantization codebook for DQ  (third) is lower when compared to VQ(fourth); lower is better. The diagonal represent the entropy for each quantization vector, lower half of the diagonal is empty.}
\label{fig:imp_marg}
 \end{figure*}
\textbf{Auto-Encoder} (AE) is an unsupervised class of DNN architectures that learns compressed feature representations from high dimensional data. Work by Kingma \etal \cite{vae} extends AE to \textit{Deep Latent Variable Models} with variants such as Variational Auto-Encoder (\textit{VAE}). For some input $x$ and a latent space $z$, VAE is composed of a decoder $p(x|z)$, a prior $p(z)$, and an encoder $q(z|x)$. VAE is a probabilistic model that implicitly learns underlying variables used to generate the data and their \textit{latent factors} by minimizing the divergence between the encoded representation $q(z|x)$ and the true data manifold $p(z)$. To evaluate AE, we can use Mutual Information (MI) which is a statistical dependence metric between two variables s.t. $I(X;Y)=H(X) - H(X|Y)$, where $H(X)$ is the \textit{information entropy} of $X$. The optimization objective of a VAE \cite{beta-vae,beta_understanding} is an upper bound to 

\begin{equation} \label{eq:1}
max[I(z; p(x|z)) - \beta I(x;z)] 
\end{equation}
that maximizes the mutual information between latent representation and decoded data, and discards information from x that is not informative to decoding $p(x|z)$. As such, maximizing \cref{eq:1} also maximizes the entropy of $z$ or ``informativeness'' \cite{hutter2002distribution}.

Our view on quantization is based on the interpretation by Richardson \etal \cite{richardson2008modern} and MacKay \etal \cite{mackay2003information}. For the sake of brevity, we refer the reader to their work for a detailed analysis and attach our own analysis and proofs in the supplementary material. 

\textbf{Scalar Quantizer} (SQ) with a vocabulary of size $K$ is an encoding function for an element $X_i$ from a sequence $X\in R^N$ of length $N$ such that $f(X)=\{1 \dots K\}^N$. SQ quantizes every element of the sequence in a \textit{memory-less} fashion with the same encoding function. SQ cannot place assumptions on cross-correlation between different sequence elements. SQ performs optimally when the probability density function (pdf) of all data is known in advance. An encoding function that follows a uniform distribution (i.e. floor function) will perform optimally when all ${X_U}_i \in X_U $ are also uniformly distributed and bounded such that ${X_U}_i \in [a,b]$. When ${X_U}_i$ has an unknown pdf, SQ will \textit{assign} probability \textit{mass} on unlikely regions in $[a,b]$.

\textbf{Vector Quantization} (VQ) ``learns'' a mapping between $X \in R^N$, and $K$ quantization vectors, or \textit{codes}. A \textbf{Codebook} is the set of \textit{codes}  $c \in R^N $ such that $C=\{c_i : i \in 1, \dots, K \}$. The VQ \textit{decoding} function returns the code $c$ with the lowest \textit{decoding error} $d$ between $c$ and the vector $X$ such that $\hat{X}=\text{VQ}(X)=c_{j_{min}}$ where  $j_{min} = argmin \{ d(X,c) : c \in C \}$. The objective function is to minimize the error of the closest codebook vector $c$ to the feature vector $X$ and can be summarized as $l_{VQ}=\underset{c \in C}{{min}}\;{d(X,\hat{X})}$. 

 \begin{figure*}[ht]
\centering
\begin{minipage}[h]{0.90\linewidth}

         \includegraphics[width=1\textwidth]{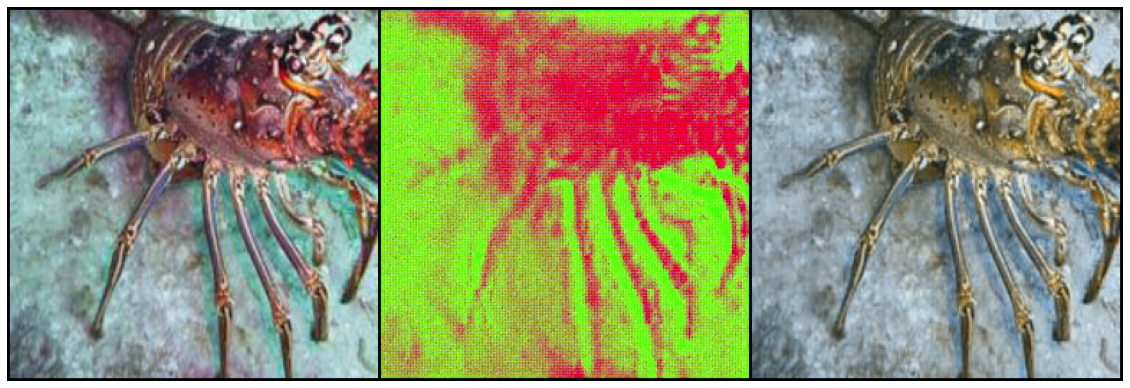}

\end{minipage}

\centering         
\caption{Original image (left) is reconstructed using only \textbf{top} level codes (middle) and only \textbf{bottom} level codes (right). Top level hierarchy contains structural information, while bottom level hierarchy contains details.}

\label{fig:hier_learning}
 \end{figure*}

\textbf{Product Quantization} (PQ) decomposes a one dimensional vector $X \in R^N $ to sub-vectors $\{X_j : j = 1, \dots, M \}$ and optimizes for a unique pair of a VQ and the sub-vector space. For $M$ different Codebooks ${C_j : j \in {1,\dots, M}}$ there is a one-to-one mapping with each $X_j$. The PQ decoding function is the concatenation or addition of all VQ decoding $\text{VQ}_j=\hat{X_j}$ for codebook $C_j$ such that $\text{PQ}(X) = {\mathbin\Vert}_{\forall j \in M} VQ_j(X_j)$. We adopt the \textit{feature decomposition} from PQ and extend it to high dimensional feature vectors to reduce the statistical independence among latent features. 

\textit{Cost} of a quantizer is the number of Codebook vectors s.t. $ C_\text{cost}=K\times M$, for PQ. \textit{Representation Capacity} ($C_R$) defines an upper bound on the sample space from the number of discrete latent factors that can be represented by the quantizer for independent random variables $X_j$, such that $S=K^M$ for $K$ codes and $M$ decomposed sub-vectors. For redundant $X_j$, the sample space is reduced to $S^{\text{new}}=({K-1})^M$ and thus the capacity is bounded by the sample space s.t. 
\begin{equation}\label{eq:cr}
C_R=-H(\textbf{X})
\end{equation}

Note that for PQ, $ C_\text{cost} $ grows linearly while $C_R$ grows exponentially, in contrast to a VQ which has linear growth for both, and thus has an exponential cost with an identical capacity to PQ. More detailed analysis and proofs can be found in the Appendix.  

\textbf{Distribution of Prior} can have an effect on the decoding performance of the quantizer. For example, VQ with $X_U$ from before can achieve identical decoding error as SQ but at a significant cost of $K^N$ as compared to $K$ for a \textit{memory-less} SQ. The assumption on the distribution of the prior can determine the cost and the representation capacity of a quantizer. 

The difference between PQ and VQ is the assumption of co-variance among features. Contrary to VQ, PQ takes advantage of the low co-variance among feature sub-vectors. 

\section{Depthwise Quantization}

Given an output feature tensor from encoder $\textbf{X} \in R$ with rank $r$, Depthwise Quantization (DQ) applies $M$ quantizers $VQ_i$ pair-wise on decomposed tensor \textit{slices} $\textbf{X}_i=\textbf{X}_i^{\alpha}$ along an axis $\alpha$ with quantization \textit{dimension} $D=|\textbf{X}_i^{\alpha}|$.
\begin{align}\label{eq:dvq}
\text{DQ}(\textbf{X}) = \{ VQ_i (\textbf{X}_i) : i = 1, \dots, M\}
\end{align}
Each $VQ_i$ optimizes Codebook $C_i$ and uses $l_{VQ}$ to define the error between $\textbf{X}_i$ and closest quantization vector $\hat{\textbf{X}}_i=Q_i(\textbf{X}_i)$. The optimization objective is the joint optimization over each codebook such that
\begin{equation}
\underset{C^1,\dots,C^M}{min}{[L_{DQ}=\sum_{\forall \hat{\textbf{X}}_i,\textbf{X}_i }{l_{VQ}(\textbf{X}_i,\hat{\textbf{X}}_i)}]}
\end{equation}
We use the $L_2$ norm as a similarity metric for $l_{VQ}$ between each $\textbf{X}_i$ and the \textit{local} quantization vector $\hat{\textbf{X}_i}$. The DQ loss is then added to the reconstruction loss of the DNN and the gradients are copied from the quantized vector $\hat{\textbf{X}_i}$ to $X$ using auto-grad \cite{stop_grad}. The loss function of DQ becomes
\begin{equation}\label{eq:loss_dnn}
    L = L_{DNN} + L_{DQ}(sg(\textbf{X}),\hat{\textbf{X}}) + \beta L_{DQ}(\textbf{X},sg(\hat{\textbf{X}}))
\end{equation}
where $sg$ stands for stop-gradient operator that stops the operand from updating during the training phase. Similar to the setting in VQ-VAE, the first loss term is used to lower the reconstruction error, the second term adjusts the codebook corresponding to the encoder output, and the third term is used to prevent the output of the encoder from growing arbitrarily. Note that the KL divergence is a constant equal to $Mlog(K)$ as DQ assumes a uniform prior distribution of latent embeddings. Therefore, the KL divergence term is dropped from the optimization objective of our framework. A detailed explanation is provided in \cref{sec:decoupled_feature_space}


Note that for a feature tensor of rank one, DQ is identical to SQ when a single codebook of dimensionality one is used. When more than one codebook is used, DQ is identical to PQ and Additive Quantization with addition as the decoding function. The advantage of DQ over other quantization methods comes from the decomposition of a tensor to sub-tensors along the axis of weak statistical dependence. 

For 2-D Convolutional Neural Networks (CNN), $\textbf{X}$ is a feature tensor of rank 3. \cref{fig:architecture} provides an illustration of the DQ process for a 3-rank tensor. Different quantizers are applied for each slice of the channel axis, but the same quantizer is applied on the sub-vectors of a feature sub-tensor, such as decomposing along the spatial dimension. 

\subsection{Decoupled Feature Space}\label{sec:decoupled_feature_space}

\textbf{Decoupled} refers to the statistical independence between features and  \textbf{Coupled} refers to the statistical dependence between features. We use \textit{Information Theory} to analyze quantization as an encoding function with an information bottleneck on a signal.

\cref{eq:1} provides the basis of the VAE optimization objective that can be formulated as a lower bound to the channel capacity as $\mathbb{L} \geq \mathbb{E}_{q({z}|{x})} \log{p({x} | {z})} - \beta D_{KL}(q({z} | {x}) || p({z}))$ \cite{beta-vae,beta_understanding} where $\beta$ is the Lagrange multiplier. $\beta$-\text{VAE} assumes a Gaussian prior $p(z) \sim \mathcal{N}(0,I)$, and DQ assumes a uniform prior. Thus the KL-Divergence of the uniform distribution and decoder is the capacity of the quantizer $D_{KL}(q(\textbf{z} | \textbf{x}) || p(\textbf{z}))=C_R$. The detailed proof can be found in Appendix 1. 
\begin{equation}\label{eq:dq_opt}
max[\mathbb{E}_{q({z}|{x})} \log{p({x} | {z})} - C_R]
\end{equation}

Reducing the capacity of the information bottleneck in VAE encourages disentangled representations in $\beta$-VAE. In a similar fashion, reducing $C_R$ encourages disentangled representations for each codebook, with the upper bound controlled by $K$ and $M$. By doing so,  significantly compressed representations can be learned for an improved downstream training objective. 

$C_R$ and by extension $H(z)$ in the discrete case are not differentiable with respect to the DNN parameters and can not be explicitly minimized. We observe \textit{index collapse} and performance degradation on hierarchical deep quantization variants. Index collapse causes the quantizer to utilize only limited number of codes. We enforce a uniform prior through an approximate solution and discuss the implications in \cref{sec:implicit_marginalization} and \cref{sec:discussion}. 

 \begin{figure*}[ht]
\centering
\begin{subfigure}{0.48\textwidth}

         \includegraphics[width=\textwidth]{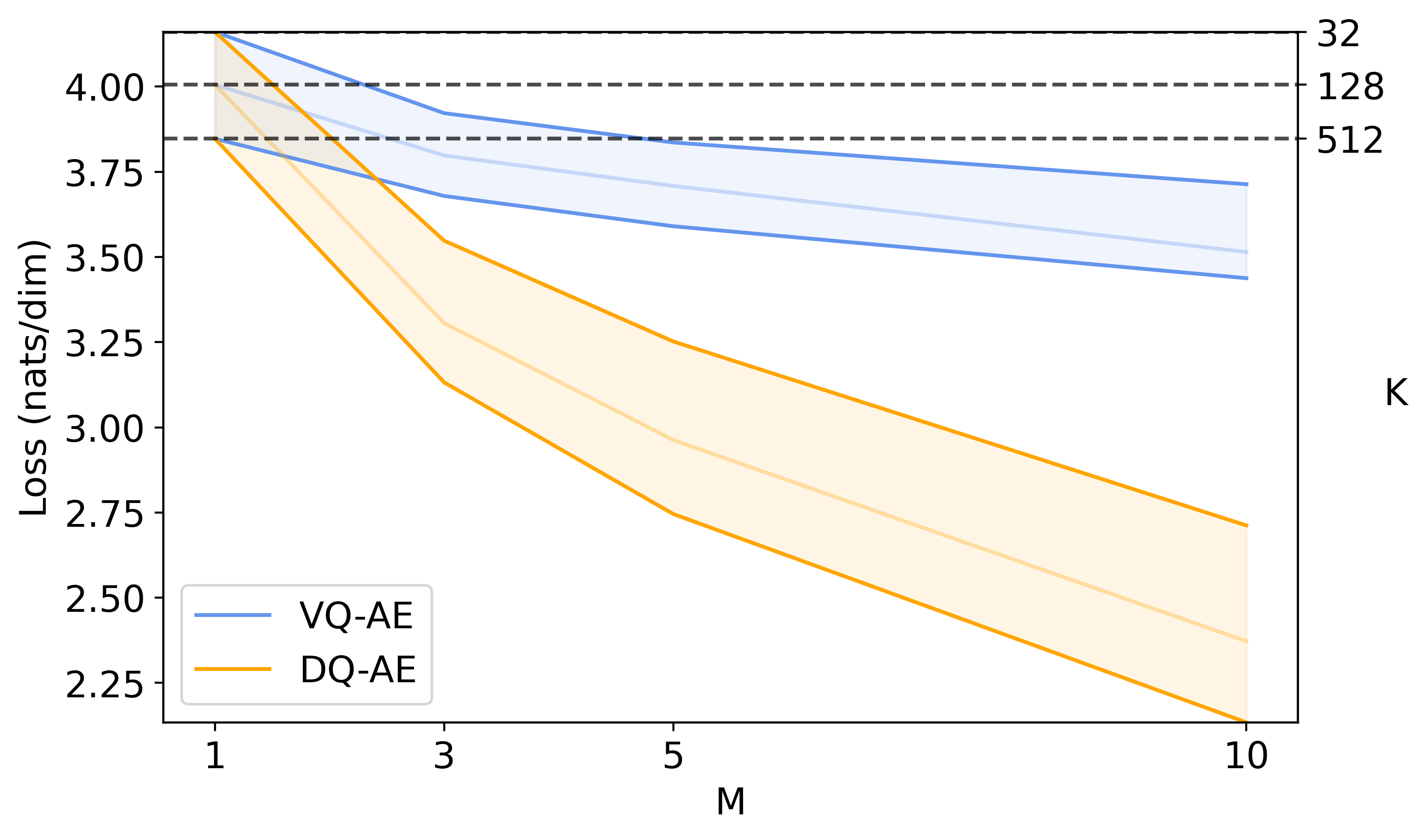}
\end{subfigure}
\begin{subfigure}{0.48\textwidth}
         \includegraphics[width=\textwidth]{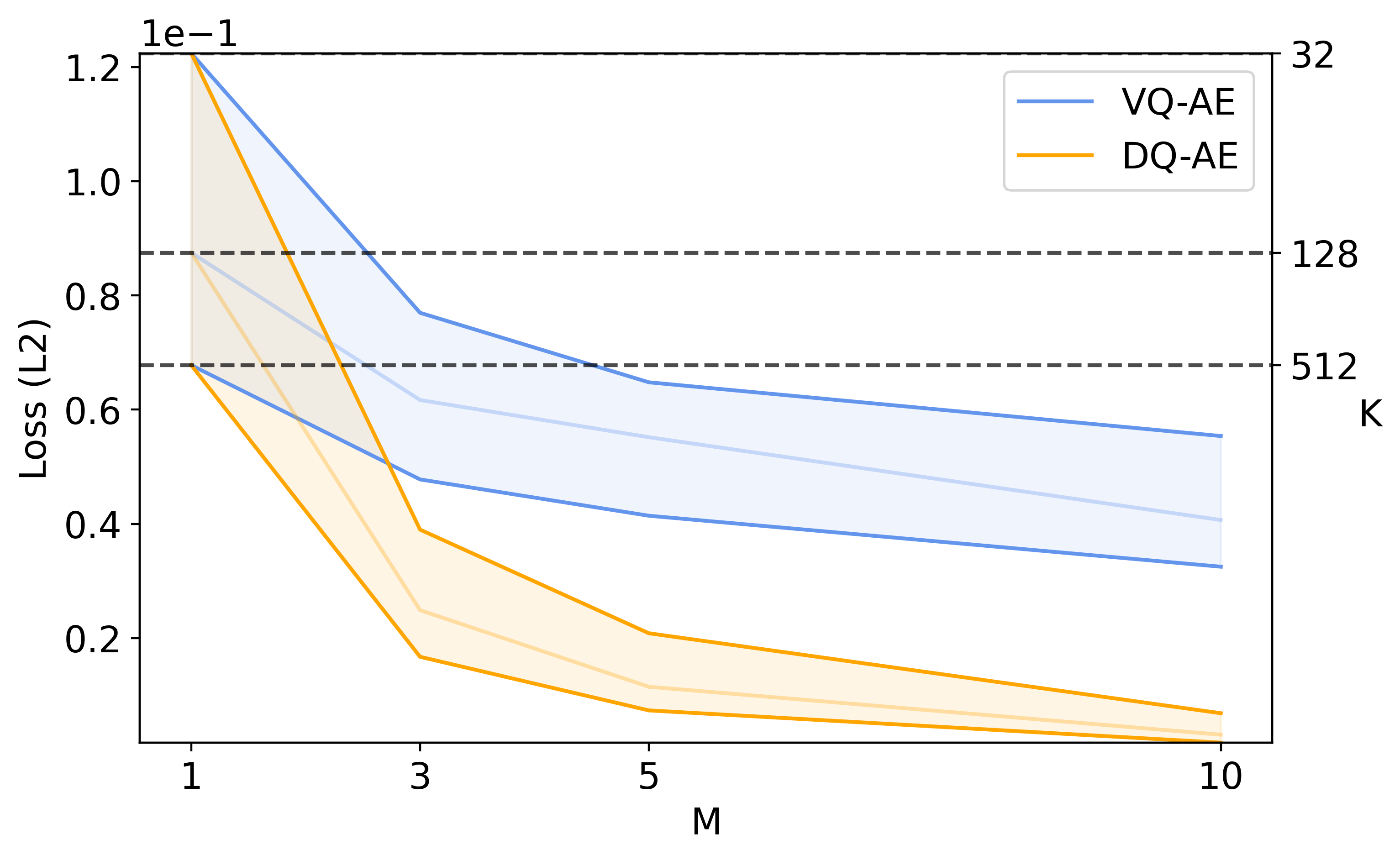}

\end{subfigure}
\centering
         \caption{Ablation study model comparison. For M=1, DQ is identical to VQ. We train models under the following settings $K=\{32,128,512\}$, $M=\{1,3,5,10\}$ and optimize for different reconstruction losses NLL (left) and $L_2$ (right). We report the average of the final convergence loss from multiple runs (10). The top bold line for each polygon corresponds to $K=32$, the bottom line to $K=512$}
\label{fig:ablation}
 \end{figure*}

\subsection{Implicit Decoupling}\label{sec:implicit_marginalization}

\textit{Implicit} decoupling of the feature space is the surrogate optimization objective derived by the explicit minimization of the decoding error. There is a joint optimization objective when DQ is applied to the intermediate feature representations in the context of DNN and trained end-to-end. DQ minimizes the decoding error along the DNN objective function. DQ works as a bottleneck on intermediate feature representations between subsequent layers of the network. We use the result from work by $\beta$-VAE on the interpretation of AE as the information bottleneck. 

$\beta$-\text{VAE} uses $q(z|x)$ to learn a set of additive channels $z_i$ where their capacity is maximized when all $z_i$ are independent. This provides an implicit optimization objective by optimizing \cref{eq:1}, which is the \textit{equivalent} in the quantized case as optimizing \cref{eq:dq_opt}. There is an equivalence between each $z_i$ and the codebook as they both perform as additive information channels that, when combined, reconstruct an original signal. Lastly, both $z_i$ and the quantizer are parametric density estimators or smaller networks that can be considered as part of a generic Network-in-Network (NiN) family of models. 

Feature independence improves downstream task performance when learned implicitly in NiN models. Xception uses Depthwise Seperable Convolution (DSC) 
to outperform \textit{coupled} variants on ablation studies on Mobile-Net \cite{intra_kernel_cor}. Additional previous analysis on the intra-kernel correlations \cite{bsconv} has demonstrated the benefits of a decoupled feature space along the channels of an image feature tensor. We corroborate the analysis with MI estimation on a static prior to determine the axis of weak statistical dependence and apply DQ along the channel dimension (``depth-wise'') and spatial dimension (``pixel-wise'') in the context of DNN.

\textbf{Uniform Prior}
In contrast to the traditional Variational Auto-Encoder, DQ relies on the assumption of uniform distribution of quantized vectors \(p(z)\). However, the assumption of a uniform prior is not strong, which can potentially lead to degrading performance and be sensitive to the random initialization. A non-uniform prior will cause \textit{codebook collapse} where only few codes are utilized in a codebook. To mitigate this issue, we follow previous work, and use Exponential Moving Average (EMA) and random re-initialization of codes. We re-initialize codes with low usage frequency counts that are below a threshold.
 Although previous works \cite{fast_decoding,on_line_em} discuss the equivalence of quantization with EMA and the $\beta$-VAE objective, there is no exact relationship between the two. VQ-VAE is an approximation to the Varitional Information Bottleneck (VIB) when trained with soft Expectation Maximization (EM). The E-step on the update rule of DQ is approximated with EMA over mini-batches of data \cite{on_line_em,chen2018stochastic}. This is in contrast to hard - EM where quantization is deterministic \cite{roy2018theory}. Soft - EM provides a probabilistic discrete information bottleneck as discussed in work by Roy \etal\cite{roy2018towards} and Wu \etal\cite{wu2018variational}. 

We use entropy of the quantization vectors to measure their information density. A successful decoupling method should generate feature vectors with high entropy. 
\textbf{Entropy Estimation} on continuous distributions is intractable, but a signal can be discretized by quantization with both parametric and non-parametric optimization on the quantization interval. The entropy is then computed on the quantized discrete distribution. 

\begin{equation}\label{eq:mi_estimate}
    {I} (X;Y)=\sum _{y\in {\mathcal {Y}}}\sum _{x\in {\mathcal {X}}}{p_{(X,Y)}(x,y)\log {\left({\frac {p_{(X,Y)}(x,y)}{p_{X}(x)\,p_{Y}(y)}}\right)}}
\end{equation}

We use the quantization regions of VQ as a density estimator for entropy, and thus mutual information on a continuous prior. When DQ is learned end-to-end, entropy can be calculated directly by the frequency count of each code vector over a sample set. Our approach in approximating MI is similar to previous work that uses Kernel Density Estimators \cite{kernel_density_estimator} and is performed post-hoc on a trained network or by training a different DQ. Quantization post-hoc is sensitive to sample size but performs at par with other state-of-the-art approaches \cite{k-means-mi,k-means-mi-2,MI_estimation,paninski2003estimation}. 

\subsection{Depth-Quantized AutoEncoder}Depthwise Quantized Auto-Encoder (DQ-AE) uses DQ at different hierarchical feature representations. The full algorithm that defines the training process is found in the supplementary material. In summary, we decode each quantized representation conditioned only on the quantized representation of the previous level. We perform this operation top-bottom and use \cref{eq:dq_opt} as the optimization objective of each DQ. Through experiments, we find that lower capacity \textit{bottom}-level hierarchy enforces the utilization of top-level hierarchies and that the problem of under-utilization of top or bottom level hierarchies can also be a consequence of over-fitting. During the early stages of training, both hierarchies are used equivalently, but at later stages, \textit{top}-level prior collapse. Our architecture leads to informative top and bottom level hierarchies as can be seen in \cref{fig:hier_learning}.

\section{Experiment}

\begin{table*}[ht]
\centering
\begin{tabular}{ c c c c c c  }
 \hline
  &\textbf{CIFAR-10} & &  \textbf{ImageNet-32} & & \textbf{ImageNet-64}  \\ 
 \hline
Model (Param.) & bits/dim & Param. & bits/dim & Param. & bits/dim   \\
S-Tr.$^1$ (59M) & 2.80 &  Img-Tr.$^2$ (-) & 3.77 & S-Tr.$^1$ (152M) & 3.44   \\
VDVAE$^3$ (39M) & 2.80 &  119M & 3.80 & 125M & 3.52  \\ 
\hline
(\textbf{Ours}) (\textbf{22M})  & \textbf{ 2.52 } &  \textbf{22M} & \textbf{  3.12} &  \textbf{22M} & \textbf{ 2.89} \\
\hline 
\end{tabular}%
\caption{Baselines: $^1$Sparse Transformer \cite{child2019generating} $^2$Image Transformer \cite{parmar2018image} 
$^3$VD-VAE\cite{vdvae}. ``Ours'' is a 2 hierarchical DQ-AE with $K$ set to 256 and 128 for top and bottom codebooks respectively. }
\label{exp:sota}
\end{table*}

In our experiment, we evaluate DQ in two settings: on a static prior, and when trained end-to-end with a DNN, on a learned prior. We first evaluate our theoretical claim on a static prior and perform an ablation study on DQ and DQ-AE. We report the details of the training and network hyper-parameters in the supplementary material. 

\subsection{Density Estimation}\label{exp:density_est}
We experimentally verify our claims on the decoupled feature space from \cref{sec:decoupled_feature_space}. The penultimate feature representation from pre-trained VGG-16 model\footnote{\url{https://pytorch.org/vision/stable/models.html}} is used on ImageNet \cite{deng2009imagenet}. DQ decomposes the feature representation ``channel-wise'' ($DQ_C$) and ``pixel-wise'' ($DQ_S$). 
The penultimate feature tensor with shape $[512 \times 7 \times 7]$ is sliced along the channel axis into 7 segments and zero padded with $D=74$ and $M=7$. The quantizers for both networks independently quantize each row for $DQ_S$ in contrast to each slice for $\text{DQ}_C$.

We train DQ as a quantizer for multiple random runs (10) and we report the mean $L_2$ norm between $\textbf{X}$ and the reconstruction $\hat{\textbf{X}}$. For each quantization method, we approximate the entropy of the codes to determine their respective information density using \cref{eq:mi_estimate}. The results of our experiments can be found in \cref{res:tab_density}.


We find that DQ can achieve better density estimation along the channel dimension as opposed to the spatial dimension. The lower entropy $H(\textbf{X})$ of the feature tensor is due to a higher redundancy among feature sub-tensor and corresponds to a higher reconstruction error. DQ can perform better when decomposing on the channel axis and our results agree with previous analyses on intra-kernel correlations \cite{bsconv}. 

\begin{table}
\begin{center} 
\begin{tabular}{ c c c c c c }
 Quant. & K & D & $L_{DQ} \downarrow$ & $\sim H(\textbf{X}) \uparrow$ \\ \hline
 Pixel  & 32 & 74 & 0.192$\pm 0.002$ &  1.98$\pm 0.01$\\  
 Channel & 32 & 74 & \textbf{0.184}$\pm 0.001$ & \textbf{2.53}$\pm 0.01$  \\  
 Pixel  & 1024 & 74 & 0.523$\pm 0.003$ & 3.64$\pm 0.01$\\  
 Channel & 1024 & 74 &\textbf{0.480}$\pm 0.001$ & \textbf{3.99}$\pm 0.01$\\  
\end{tabular}
\caption{Density estimation on ImageNet feature space extracted from VGG-16. Results are from 10 train runs with random code initialization. DQ applied along the channels (Channel Quantization) as opposed to the spatial dimension (Pixel Quantization). $D$ is the size of the feature vector, $K$ the discrete codes used, $L_2$ the reconstruction error, and $H(\textbf{X})$ is the mean entropy of the feature tensor. Channel Quantization surpasses Pixel Quantization in all respects. }
\label{res:tab_density}
\end{center}
\end{table}

\subsection{Implicit Decoupling}\label{exp:marg}

We train a DQ-AE and a VQ-VAE \cite{vqvae} for $M=10$ and $K=512$ with an identical network configuration, methodology and hyper-parameters. 
The difference between architectures is highlighted in \cref{fig:architecture}. We measure the likelihood estimation of the two approaches on CIFAR-10\cite{krizhevsky2009learning} and quantize each image to $8\times8\times10$ codes. VQ-VAE NLL is \textbf{4.36} bits/dim as compared to \textbf{3.14} bits/dim for single hierarchy DQ-AE, a 28\% decrease. 

\textbf{High Entropy} We show that the learned features of DQ have high entropy which indicates low statistical dependence among them. In contrast, VQ appears to have few very informative features and many uninformative ones. The mean entropy of the prior is $ H(z)=6.03$ nats/pixel for DQ as compared to  $H(z)=5.86$ nats/pixel for VQ. The entropy distribution among spatial features of the prior can be found in the left two sub-figures in \cref{fig:imp_marg}.

\textbf{Low MI} We estimate the pairwise MI of the quantization vectors along the depth of the feature tensor with mean score of $1.93$ and $2.36$ nats/vector respectively. A comparison matrix can be found in the right two sub-figures in \cref{fig:imp_marg}. For DQ the MI between quantization vectors is significantly lower as visualized by the mostly empty upper triangular matrix. In contrast, for VQ there seems to be higher redundancy among quantization vectors. The diagonal of the matrix represents the entropy of each quantization vector. The MI estimate on the quantization vector shows that the redundancies are significantly higher in the learned representation for VQ.

\subsection{Ablation study}\label{exp:ablation}
We study the effect of $K$ and $D$ on the model performance. The model is more sensitive to the dimensionality $D$ of the sub-vector and less sensitive to $K$. DQ outperforms coupled variants on likelihood estimation and reconstruction loss in all settings. $C_R$ grows exponentially with $M$ as opposed to $K$. Fewer code vectors can be used to quantize without performance degradation. For example, when $M=3$, DQ outperforms a VQ variant by 35\% and uses 25\% fewer code vectors. \cref{fig:ablation} shows a summary of the loss for different $K$ and $M$ values. A detailed table of the results can be found in the Appendix.  


\subsection{Likelihood Estimation}\label{exp:likelihood}

For likelihood estimation, we compare DQ-AE with other likelihood estimator models and report the numbers from their work. We use Very Deep VAE (``VD-VAE'') \cite{vdvae} as a continuous AutoEncoder baseline and Sparse Transformer  (``S-Tr'') \cite{child2019generating} as an Auto-Regressive baseline. 

For experiments on ImageNet, we add a number on the image resolution at which we train the model at the end of the dataset name. For our model, we use an identical architecture and number of hierarchies for all resolution of the dataset. The detailed results are in \cref{exp:sota}.  
We outperform all previous state-of-the-art models when measuring the loss in bits/dim, we also report $C_R$ separately. The estimate for $C_R \sim 0.2$ nats. Visual inspection of both top and bottom hierarchies confirm that they encode different granularity of features and are utilized (\cref{fig:hier_learning}), and perceptual quality is improved (\cref{fig:comparison}). Additional high resolution images are attached in the supplementary materials. 

When compared to the hierarchical model by Razavi \etal \cite{vqvae2}, DQ-AE also outperforms in reconstruction error for $L_2$ on ImageNet-256. On CIFAR-10, the DQ-AE loss is $0.019$ compared to $0.044$ for VQ-VAE. For ImageNet-256, DQ-AE loss is $0.0032$ compared to $0.005$ for VQ-VAE-2. 

 



\section{Discussion}
\label{sec:discussion}
We thoroughly evaluate the theoretical claims of our work and empirically verify our method in likelihood estimation. Evaluation of the discrete representation on a downstream task such as latent interpolations is domain specific. Sampling from the multi-resolution and high dimensional discrete codebooks requires training additional models post-hoc. As such, there are multiple open problems in how to design such a model. We leave this for future work.

The direct evaluation and comparison on NLL between explicit likelihood models can be non-equivalent. Our model makes different assumptions on the prior distribution and as such the direct comparison can be flawed. Previous work \cite{alemi2018fixing} has suggested that the ELBO might be a poor metric to evaluate deep latent variable models. We mitigated the issue and followed the theoretical result and experimental methodology to previous work \cite{vdvae}. We consider the proper evaluation of the discrete prior with other model variants as an open problem. 

\section{Conclusion}

We analyze the effects of decomposing an image feature tensor along an axis of statistical independence. Decomposition and quantization among independent features outperforms coupled feature variants. Our theoretical insights focus on feature decoupling for decomposed image feature tensors along the channel axis. Our results corroborate previous analyses and explain the advantage of NiN applications which can be interpreted as an information bottleneck. 

Based on our theoretical insight, we propose Depthwise Quantization (DQ) that provides significantly more efficient bottleneck capacity by eliminating redundancies implicitly in the feature axis. DQ is trained end-to-end with a Hierarchical Auto-Encoder (DQ-AE) and learns improved hierarchical discrete representations. Our method is domain agnostic, and we consider the evaluation on a specific task for future work. 

\clearpage
\newpage
{\small
\bibliographystyle{ieee_fullname}
\bibliography{bib_main}
}
\clearpage
\newpage

\appendix

\section{Appendix}

\subsection{Representation Capacity}

Proofs in this section correspond to claims and results in the main text. Where applicable, a proposition will refer to the the equation in the main text for which the result is applied.

\proposition[Depthwise Quantization Channel Capacity - Result for Equation \noref{eq:cr} in the main text]{
The capacity $\mathbfcal{C}$ of Depthwise Quantization (DQ) channel for set of codebooks $C$ is the entropy of the codebooks s.t. $\mathbfcal{C}=H(C)$

}


\begin{proof}
  Let the capacity of a channel $\mathbfcal{C}=I(x;z)$ \cite{tishby2000information} , where $I(:;:)$ is the mutual information. It is sufficient to show $\mathbfcal{C}=I(x;C)=H(C)-H(x|C)$ where $z=C=\{ C_i : i \in N\}$ is the set of codebooks. Since the quantization channel is a noiseless discrete channel with deterministic quantization function, $P(x|C)=1$ and thus $H(x|C)=0$. \qedhere

\end{proof}

\begin{equation}
\mathbfcal{C}=I(x;C)=H(C)
\end{equation}


\proposition[Representation Capacity - Result for Equation \noref{eq:cr} in the main text]{
The channel capacity is bounded by the number of discrete latent factors $S$ that can be represented by DQ. Let $N$ be the cardinality of the set of codebooks $C$ with $K$ codes. \textit{Representation Capacity} is defined as $C_R=-H(C)=\log{S}$
}

\begin{proof}

Let $S=K^N$ be the sample space for the set of codebooks $C={C_i : i\in N}$ with $K$ codes. By definition \begin{equation}H(C)=-\sum_{C_i \in N}{P(C_1,...,C_n)\log{P(C_1,...,C_n)}}\end{equation} where $P(C_i)P(C_j)>P(C_i)P(C_j|C_i)$.\\
$H(C)$ is maximized when $C_i,C_j$ are independent variables and are uniformly distributed (uniform prior) s.t. $P(C_i)=\frac{1}{K}$. Thus:
\begin{align*}
H_\text{max}(C)&=-\sum_{i \in K}[P(C_1)\times ...\times P(C_n)\\ \nonumber
&\log{[P(C_1)\times ...\times P(C_n)}] \\ \nonumber
&=\log{K^N}=\log{S} \nonumber \qedhere
\end{align*}
\end{proof}

\begin{equation}
    C_R=-H(C)=\log{S}
\end{equation}

\proposition[ELBO for Depthwise AutoEncoder - Result for Equation \noref{eq:dq_opt}]{

The variational lower bound of DQ-AE is 
\begin{equation}\mathbb{L}\geq\text{max}[\mathbb{E}_{q({z}|{x})} \log{p({x} | {z})} - C_R]\end{equation}
}

\begin{proof}

By definition \cite{beta_understanding} \begin{equation}\mathbb{L} \geq \mathbb{E}_{q({z}|{x})} \log{p({x} | {z})} - \beta D_{KL}(q({z} | {x}) || p({z}))\end{equation}
Thus, it is sufficient to show that $C_R$ is the bound of the divergence of the uniform prior $p(z)$ and inferred prior $q(z|x)$ s.t.

\begin{equation}
D_{KL}(q(z|x)||p(z))=C_R-S
\end{equation}

  Let $p(z)$ be the uniform distribution and $q(z|x)$ the inferred prior. Therefore,
  \begin{align*}
    D_{KL}(q(z|x)||p(z)) &=\sum_{i \in N} q(z_i|x)\log \left(\frac{q(z_i|x)}{p(z_i)}\right) \\
    &=\sum_{i \in N} q(z|x)\log\left({q(z|x)}{K^{-1}}\right) \\
    &=\sum_{i \in N} q(z|x)\log\left({q(z|x)}\right) -N\log\left(K\right)\\ 
    &< -H(q(z|x))\qedhere
  \end{align*}
  
\end{proof}
\normalfont
Since $S$ is constant, it does not affect the optimization objective, the ELBO is

\begin{equation}
\mathbb{L} \geq \text{max}[\mathbb{E}_{q({z}|{x})} \log{p({x} | {z})} - C_R]
\end{equation}
\normalfont
\clearpage
\newpage
\subsection{Architecture}
In this section we provide details on the Hierarchical DQ-AE architecture. 

\begin{figure}[ht]

\setcounter{figure}{0}   

\algnewcommand{\LineComment}[1]{\State \(\triangleright\) #1}
\renewcommand\figurename{Algorithm}
\begin{algorithm}[H]
\begin{algorithmic}

 \State {\textbf{given} encoder E,  decoder $\mathbfcal{D}$, N $\times$ \{ quantizers Q, decoders D, up-samplers U \} for each hierarchy, Reconstruction Loss function $\mathcal{L}$ and Optimizer $\mathcal{O}$ and training sample x}
\\
 \LineComment{Stack of N encoded representations bottom to top}
 \State  $\textbf{e}_\text{all}\leftarrow$ E(x) 
 \State $\text{e}_{\text{top}} \leftarrow$ pop($\textbf{e}_\text{all}$)
 \\
 \LineComment{Quantize using DVQ}
 \State q $\leftarrow$ $\text{Q}_\text{top}$($\text{e}_{\text{top}}$) 
 \State d $\leftarrow$ $\text{D}_\text{top}$(q)
 \State $\text{u}_\text{all}$ $\leftarrow$ list()
\For{e in $\textbf{e}_\text{all}$}
  \State  q,d,u  $\leftarrow$ \textsc{Decode}(e, d)
 \State $\text{u}_\text{all}$ $\leftarrow$ append(u)
  
\EndFor

 \State  $\hat{x} \leftarrow $ $\mathbfcal{D}$ ( $\text{u}_\text{all}$ )
\State Update $\theta_{[E,Q,D,U]}$ based on $\mathcal{L}(x,\hat{x})$, using Optimizer $\mathcal{O}$
\\

\Procedure{Decode}{$\text{e}_\text{cur}$, $\text{d}_\text{prev}$}
 \State {\textbf{Input} Current level encoding $\text{e}_\text{cur}$} and previous decoding $\text{d}_\text{prev}$ 
 \State {\textbf{Output} Current Level quantization q, upsampling u and decoding d}
 \\
\State $\text{q} \leftarrow \text{Q}_\text{cur}$ ($\text{e}_\text{cur}$, $\text{d}_\text{prev}$) 
\State u $\leftarrow$ $\text{U}_\text{prev}$($\text{q}$)
\State $\text{d} \leftarrow \text{D}_\text{cur}$($\text{q}$) + $\text{d}_\text{prev}$ 

\Return {$\text{q}$, $\text{d}$, $\text{u}$}
\EndProcedure
 \end{algorithmic}
 \caption{N-Hierarchical Depthwise Vector Quantizer}\label{alg:n_hier}

\end{algorithm}
\caption{ As opposed to VQ-VAE \cite{vqvae2} we use skip connections on the decoded quantized representations from top hierarchies to bottom and thus increase interaction between hierarchies to avoid prior collapse of top-level hierarchies. The decoder accepts quantized upsampled representations as opposed to independently decoding each hierarchy. \cref{fig:app_arch} shows an overview of the architecture. }
\end{figure}
\setcounter{figure}{0}    

\begin{figure}[ht]
     \centering
     \begin{subfigure}[b]{\linewidth}
         \centering
         \includegraphics[width=\textwidth]{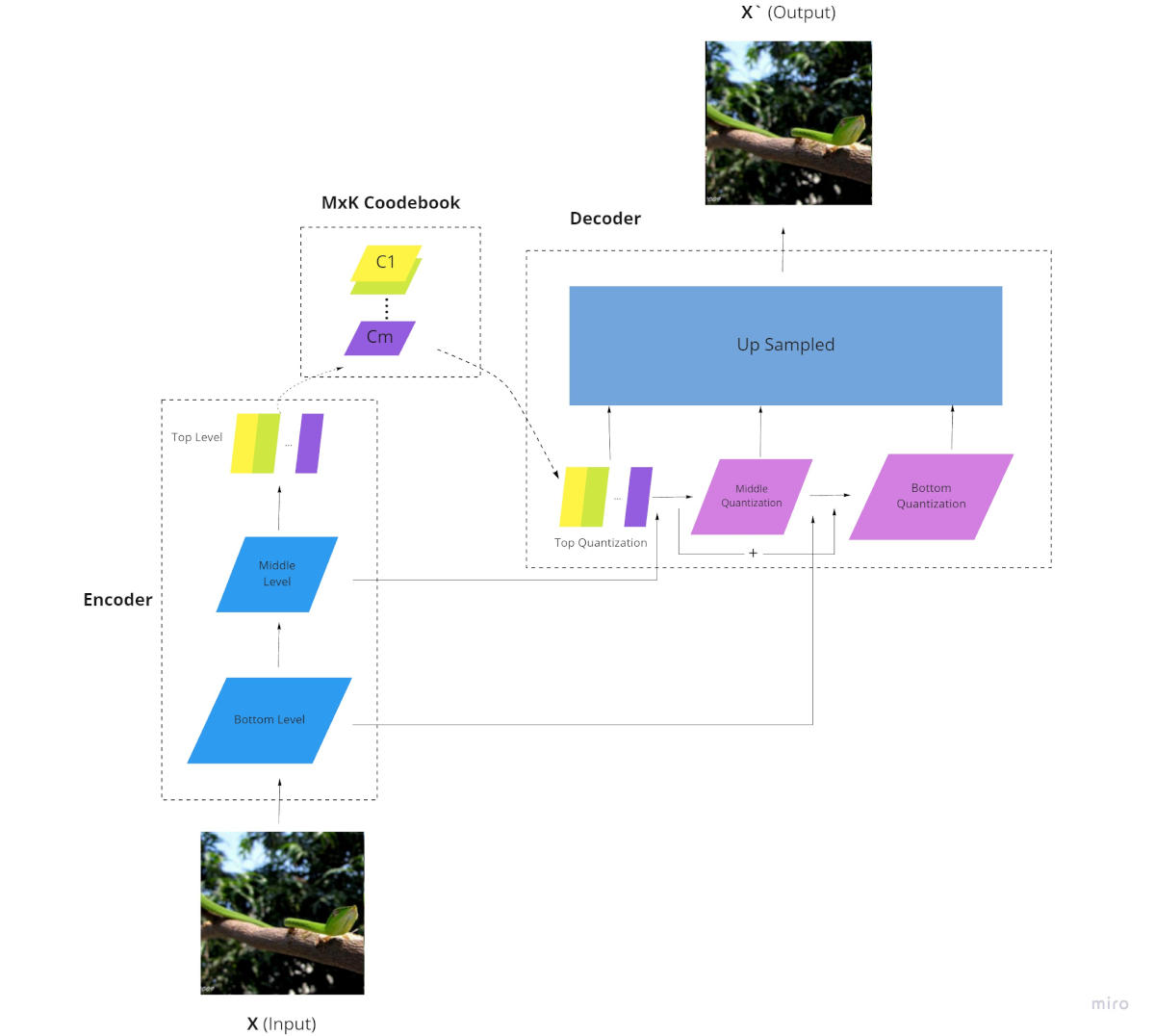}

     \end{subfigure}
     \hfill
     
         \caption{Architecture of N-Hierarchical Depthwise AutoEncoder. $X$ is input to the model and is progressively encoded to finer grain representations. Each hidden representation in the decoder is decoded using previous hierarchy's decoded quantized representation as well as the encoded representation. The quantized representations are up-sampled and decoded jointly. Quantization of top use no prior decoding. }
         \label{fig:app_arch}
\end{figure}

\begin{figure}[ht]
     \centering
     \begin{subfigure}[b]{\linewidth}
         \centering
         \includegraphics[width=\textwidth]{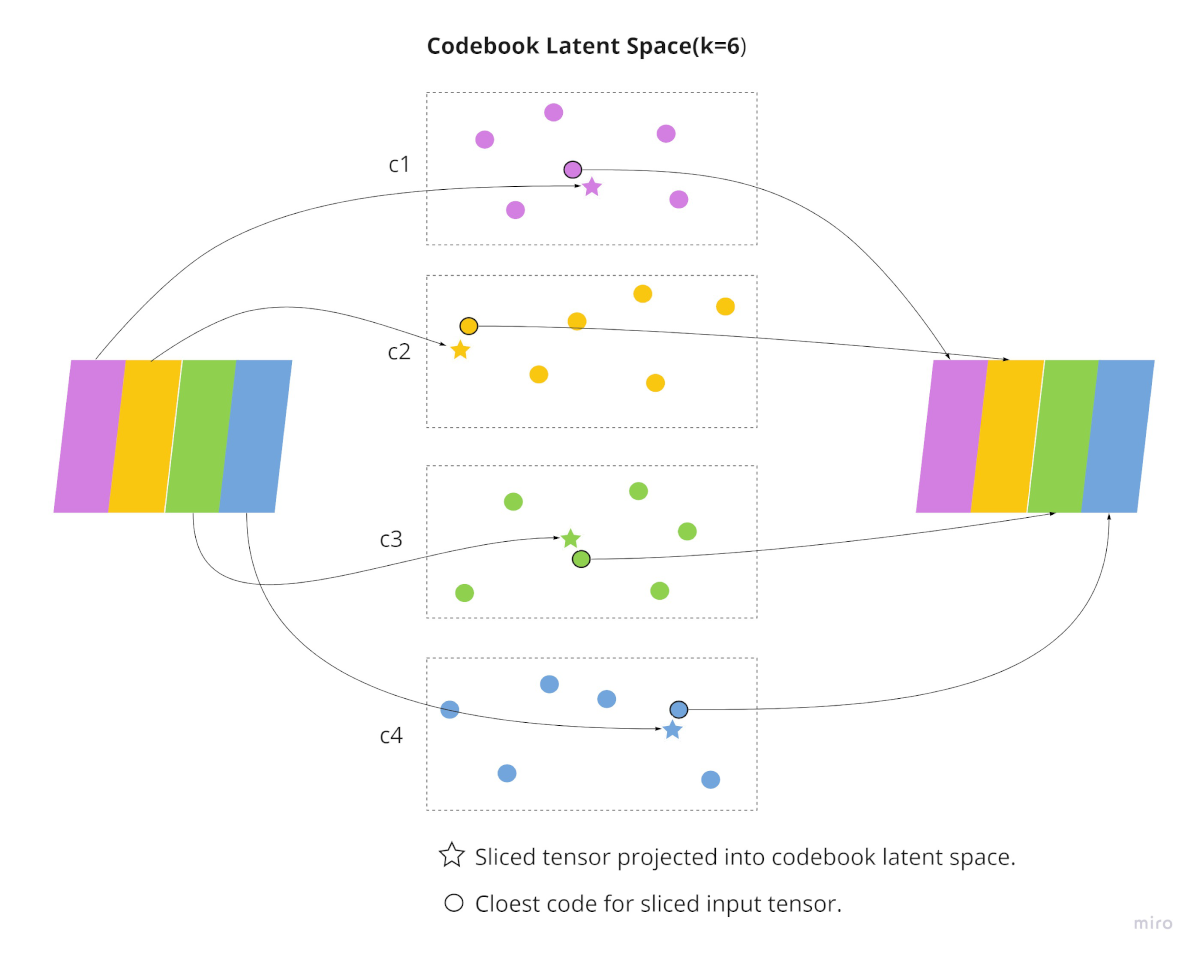}

     \end{subfigure}
     \hfill
     
         \caption{Each input latent representation is sent to the corresponding codebook. The closest code in the codebook latent space is the output of $DQ$.}
         \label{fig:quantization}
\end{figure}

\clearpage
\pagebreak
\subsection{Ablation Study} 
Results for the ablation study on the quantization process can be found in \cref{tab:loss_fn_ablt}. Results for the ablation study on DQ-AE can be found in \cref{tab:arch_ablt}. We also perform additional experiments on MNIST 
where DQ (``Our'') outperforms VQ with 1.92e-04 in $l_2$ reconstruction loss as compared to 3.41e-04, and similarly for CelebA 
with 9.57e-03 compared to 3.70e-02. 

\subsection{Training Configuration}

For all experiments and for the quantizer we use $\beta=0.25$ and dimensionality of each code $D=64$, decay factor $\gamma=0.99$ and $\epsilon=1.00e-05$ unless otherwise noted. We use a different random seed for all experiments and for every trial. For the discretized logistic mixture loss (``mix'') \cite{salimans2017pixelcnn++}, we use 10 components and discretize on 8-bit (lossless). We use Adam with weight decay regularization \cite{loshchilov2017decoupled} for optimization for all training settings. We use automatic mixed precision (\textit{amp})\footnote{\url{https://pytorch.org/docs/stable/amp.html}}.We use a batch size of 128, learning rate 2.00e-04 and train for 400 epochs.

\textbf{Ablation Study} For DQ-AE we use 2 Encoder Block composed of 4 Resnet Block with Conv2D layer of 256 channel and 256 hidden unit and stride 2.

\textbf{Likelihood estimation} DQ-AE for the likelihood estimation task uses 2 hierarchies with $K_\text{bot}=128$ and $K_\text{top}=256$. For each hierarchical encoder, it uses 2 encoder block composed of 4 resnet block with Conv2D layer of 256 channel and 256 hidden unit. 

\begin{table}[ht]
\begin{tabular}{|l|l|l|l|l|}

\hline loss func.     & M & K & DQ (nats/dim) & VQ (nats/dim) \\ \hline
ce & 1 & 32 & {4.16e+00} & 4.16e+00 \\
ce & 1 & 128 & {4.01e+00} & 4.01e+00 \\
ce & 1 & 512 & {3.85e+00} & 3.85e+00 \\
ce & 3 & 32 & \textbf{3.55e+00} & 3.92e+00 \\
ce & 3 & 128 & \textbf{3.31e+00} & 3.80e+00 \\
ce & 3 & 512 & \textbf{3.13e+00} & 3.68e+00 \\
ce & 5 & 32 & \textbf{3.25e+00} & 3.84e+00 \\
ce & 5 & 128 & \textbf{2.96e+00} & 3.71e+00 \\
ce & 5 & 512 & \textbf{2.75e+00} & 3.59e+00 \\
ce & 10 & 32 & \textbf{2.71e+00} & 3.71e+00 \\
ce & 10 & 128 & \textbf{2.37e+00} & 3.51e+00 \\
ce & 10 & 512 & \textbf{2.13e+00} & 3.44e+00 \\
\hline loss func.      & M & K  & DQ ($L_2$) & VQ ($L_2$)  \\
\hline
mse & 1 & 32 & {1.22e-01} & 1.22e-01 \\
mse & 1 & 128 & {8.75e-02} & 8.75e-02 \\
mse & 1 & 512 & {6.78e-02} & 6.78e-02 \\
mse & 3 & 32 & {3.90e-02} & 7.70e-02 \\
mse & 3 & 128 & \textbf{2.49e-02} & 6.17e-02 \\
mse & 3 & 512 & \textbf{1.67e-02} & 4.78e-02 \\
mse & 5 & 32 & \textbf{2.08e-02} & 6.48e-02 \\
mse & 5 & 128 & \textbf{1.15e-02} & 5.52e-02 \\
mse & 5 & 512 & \textbf{7.33e-03} & 4.14e-02 \\
mse & 10 & 32 & \textbf{6.84e-03} & 5.54e-02 \\
mse & 10 & 128 & \textbf{3.08e-03} & 4.07e-02 \\
mse & 10 & 512 & \textbf{1.68e-03} & 3.25e-02 \\
\hline

\end{tabular}

                      \caption{We vary the number of codebook vectors $K$ and codebooks $M$, while we keep the same $D=64$. We evaluate our results on CIFAR10 using an identical training configuration between all models and multiple random initialization. Note that the DQ model do not fully converge, due to the limited number of computational resources. We train for 400 epochs and pick the best test loss for each architecture. The comparison between the models shows a statistical trend of improved likelihood estimation for $DQ-AE$. Figure \noref{fig:ablation} in the main text, shows the aggregate results of the likelihood estimation. The top, middle, and bottom line correspond to K having values 32,128,and 512, respectively. The effect of K is not as significant as the effect of M. For M=1 both VQ and DVQ are identical in terms of theoretical and experimental performance. As we increase M, we find that the loss significantly decreases. Moreover, K, is not the limiting factor to the channel capacity but M is. This can also be seen on the graph as the loss for all different K converges as we increase M. }
\label{tab:loss_fn_ablt}
\end{table}

\begin{table*}[ht]
\centering
\begin{tabular}{|l|l|l|l|l|}
\hline loss func.    & M & K & DQ (nats/dim) & VQ (nats/dim) \\ 
\hline
ce & 5 & [128,128,128] & \textbf{2.96e+00} & 3.73e+00 \\
ce & 5 & [128,128] & \textbf{2.95e+00} & 3.60e+00 \\
ce & 5 & [128,256] & \textbf{2.96e+00} & 3.69e+00 \\
ce & 5 & [128,32] & \textbf{2.94e+00} & 3.70e+00 \\
ce & 5 & [256,128] & \textbf{2.84e+00} & 3.63e+00 \\
ce & 5 & [256,256] & \textbf{2.85e+00} & 3.63e+00 \\
ce & 5 & [32,128] & \textbf{3.21e+00} & 3.72e+00 \\
ce & 5 & [32,32,32] & \textbf{3.24e+00} & 3.69e+00 \\
ce & 5 & [64,64,64] & \textbf{3.08e+00} & 3.79e+00 \\
\hline loss func.   & M & K  & DQ (nats/dim) & VQ (nats/dim)  \\
\hline
mix & 5 & [128,128,128] & \textbf{2.55e+00} & 3.04e+00 \\
mix & 5 & [128,128] & \textbf{2.56e+00} & 3.12e+00 \\
mix & 5 & [128,256] & \textbf{2.52e+00} & 3.11e+00 \\
mix & 5 & [128,32] & \textbf{2.55e+00} & 3.18e+00 \\
mix & 5 & [256,128] & \textbf{2.49e+00} & 3.15e+00 \\
mix & 5 & [256,256] & \textbf{2.49e+00} & 3.11e+00 \\
mix & 5 & [32,128] & \textbf{2.79e+00} & 3.26e+00 \\
mix & 5 & [32,32,32] & \textbf{2.80e+00} & 3.26e+00 \\
mix & 5 & [64,64,64] & \textbf{2.65e+00} & 3.06e+00 \\
\hline loss func.      & M & K  & DQ ($L_2$) & VQ ($L_2$)  \\
\hline
mse & 5 & [128,128,128] & \textbf{1.15e-02} & 5.53e-02 \\
mse & 5 & [128,128] & \textbf{1.24e-02} & 5.05e-02 \\
mse & 5 & [128,256] & \textbf{1.02e-02} & 5.52e-02 \\
mse & 5 & [128,32] & \textbf{1.15e-02} & 5.07e-02 \\
mse & 5 & [256,128] & \textbf{9.31e-03} & 4.52e-02 \\
mse & 5 & [256,256] & \textbf{9.27e-03} & 5.13e-02 \\
mse & 5 & [32,128] & \textbf{1.94e-02} & 5.82e-02 \\
mse & 5 & [32,32,32] & \textbf{2.03e-02} & 6.29e-02 \\
mse & 5 & [64,64,64] & \textbf{1.50e-02} & 5.81e-02 \\
\hline
\end{tabular}

                      \caption{Hierarchical Depthwise Quantizers for 2 and 3 hierarchies. DQ outperforms equivalent VQ. The ``mix'' objective function refers to 8-bit mixture of logistics \cite{salimans2017pixelcnn++} following the methodology by Child 
                      \etal \cite{vdvae}. The hierarchy capacity $K$ is reported from top to bottom, i.e. $[K_{top}, K_{mid}, K_{bot}]$. }
\label{tab:arch_ablt}
\end{table*}

\begin{figure}[ht]
     \centering
     \begin{subfigure}[b]{\linewidth}
         \centering
         \includegraphics[width=.8\textwidth]{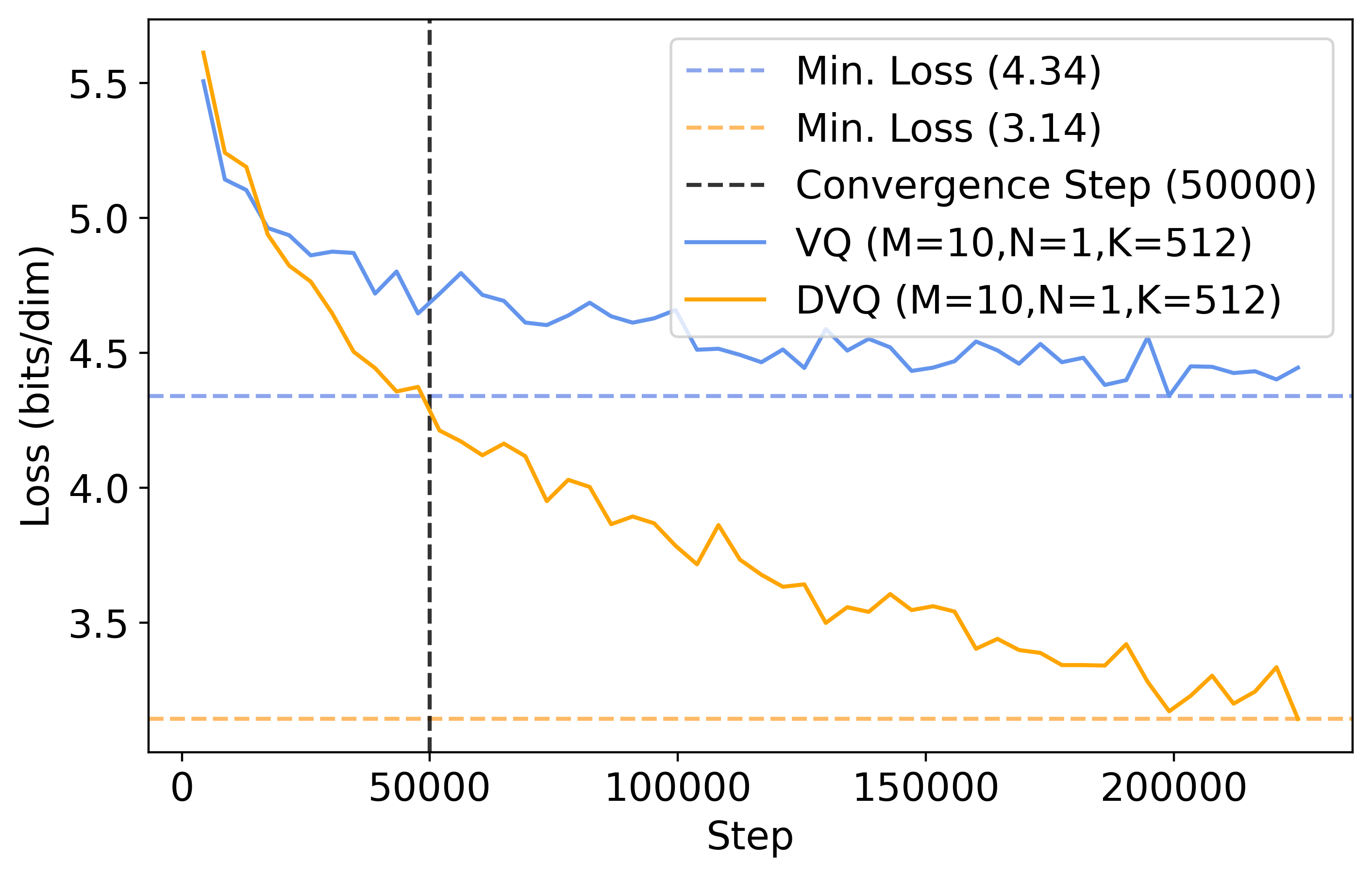}

     \end{subfigure}
     \hfill
     
         \caption{NLL Loss in bits/dim over time. Comparison between VQ and DVQ with an equivalent training set up. DVQ matches the best NLL reported for VQ by step 50,000 in contrast to step 200,000.}
         \label{fig:loss_convergence}
\end{figure}

\newpage
\pagebreak
\clearpage

\begin{figure*}[ht]

\subsection{Hierarchical Reconstruction}
     \centering
     
     \begin{subfigure}[b]{.7\linewidth}
         \centering
         \includegraphics[width=.9\textwidth]{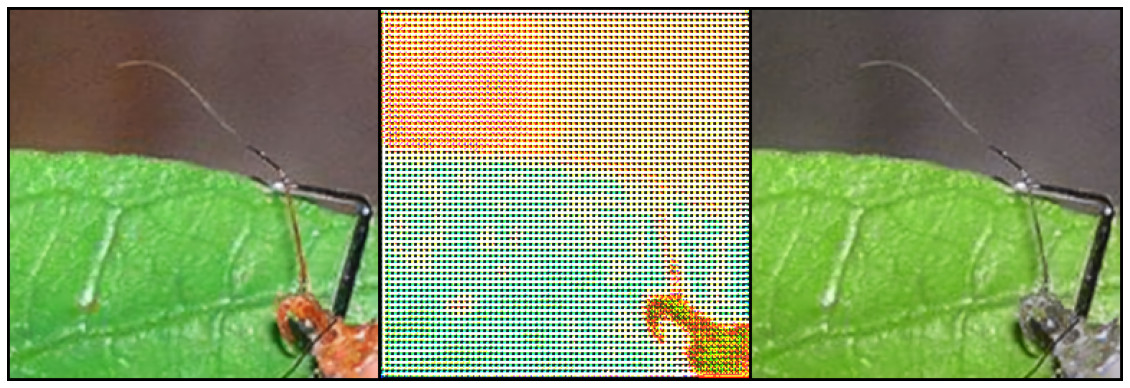}

     \end{subfigure}
     \begin{subfigure}[b]{.7\linewidth}
         \centering
         \includegraphics[width=.9\textwidth]{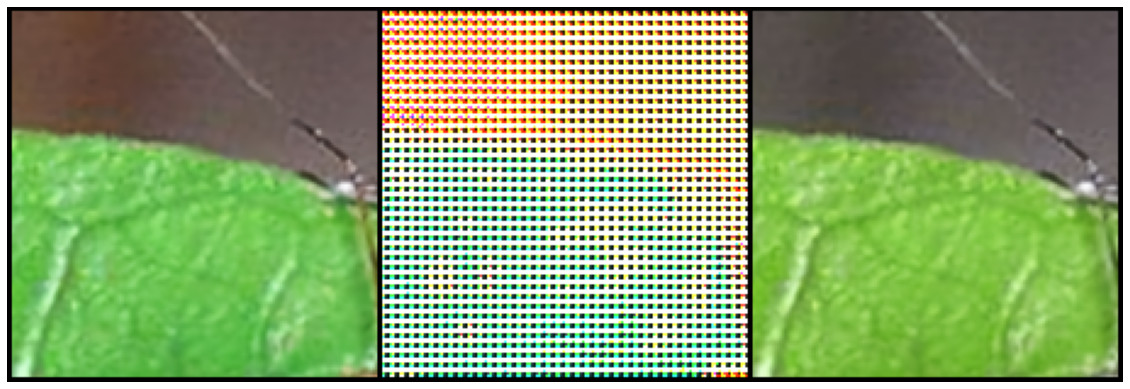}

     \end{subfigure}


     
         \begin{subfigure}[b]{.7\linewidth}
         \centering
         \includegraphics[width=.9\textwidth]{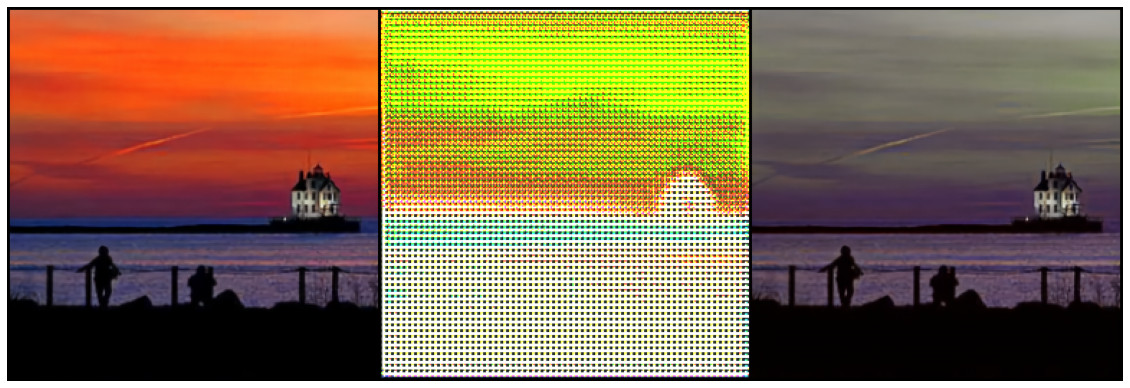}

     \end{subfigure}
     \begin{subfigure}[b]{.7\linewidth}
         \centering
         \includegraphics[width=.9\textwidth]{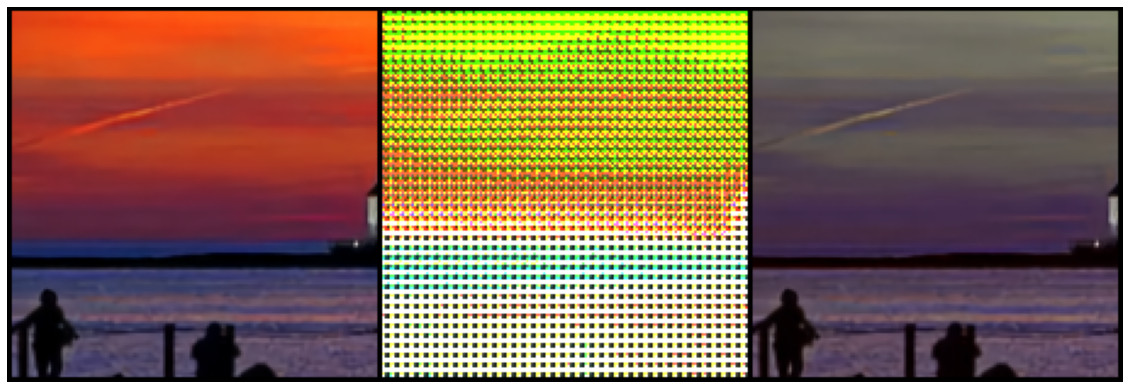}

     \end{subfigure}
                  \caption{Image reconstructions from a model trained with $L_2$ for the reconstruction loss. Original image (left) is reconstructed using only \textbf{top} level codes (middle) and only \textbf{bottom} level codes (right). Top level hierarchy contains structural information, while bottom level hierarchy contains details. }
\end{figure*}
\newpage
\begin{figure*}[ht]
     \centering
     
     \begin{subfigure}[b]{.8\linewidth}
         \centering
         \includegraphics[width=1\textwidth]{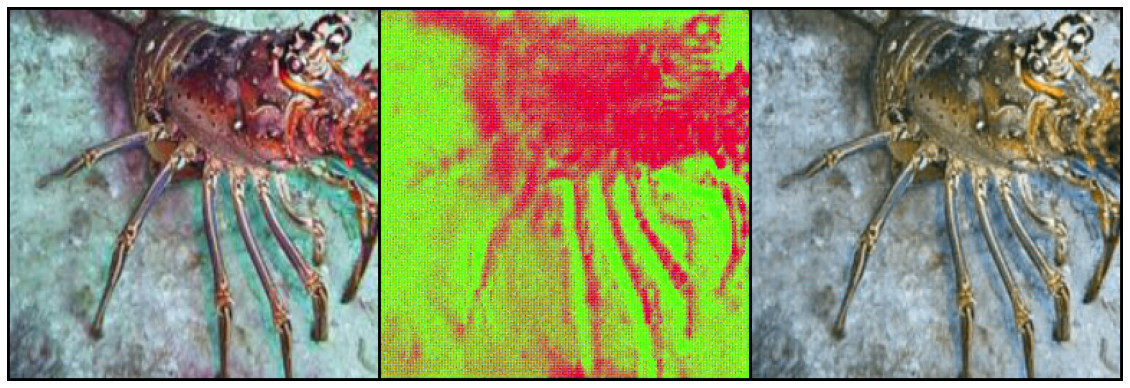}

     \end{subfigure}
     \begin{subfigure}[b]{.8\linewidth}
         \centering
         \includegraphics[width=1\textwidth]{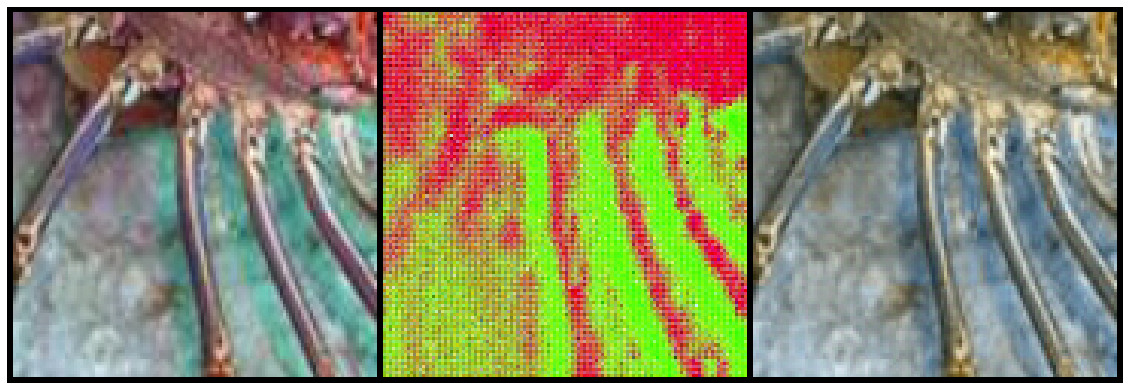}

     \end{subfigure}


     
         \begin{subfigure}[b]{.8\linewidth}
         \centering
         \includegraphics[width=1\textwidth]{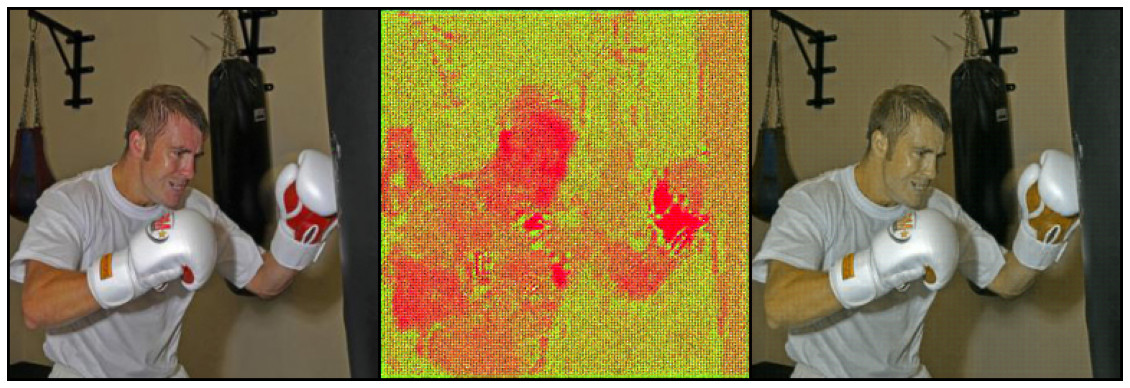}

     \end{subfigure}
     \begin{subfigure}[b]{.8\linewidth}
         \centering
         \includegraphics[width=1\textwidth]{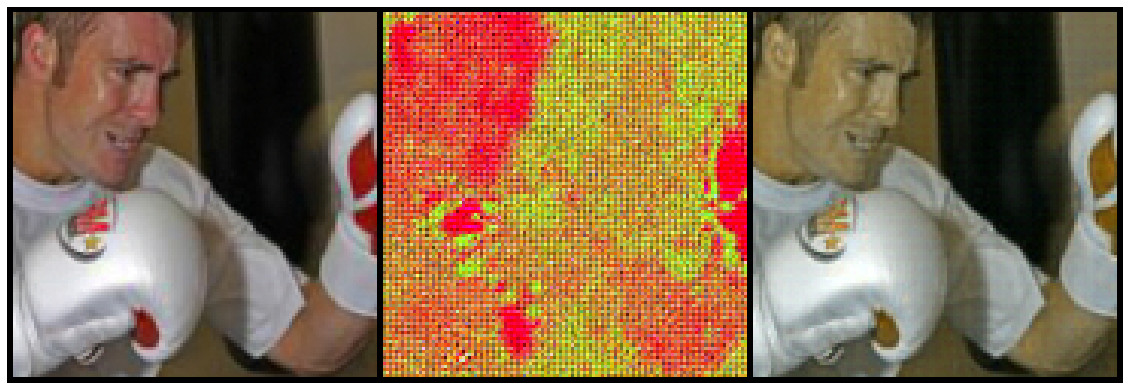}

     \end{subfigure}
                  \caption{Image reconstructions from a model trained with discretized mixture of logistic loss (dmol) \cite{salimans2017pixelcnn++} for the reconstruction loss. Original image (left) is reconstructed using only \textbf{top} level codes (middle) and only \textbf{bottom} level codes (right). Top level hierarchy contains structural information, while bottom level hierarchy contains details.}
\end{figure*}
\newpage
\pagebreak
\clearpage

\begin{figure*}[ht]
     \centering
     
\subsection{Perceptual Evaluation of Image Reconstructions}
     \begin{subfigure}[b]{.8\linewidth}
         \centering
         \includegraphics[width=.9\textwidth]{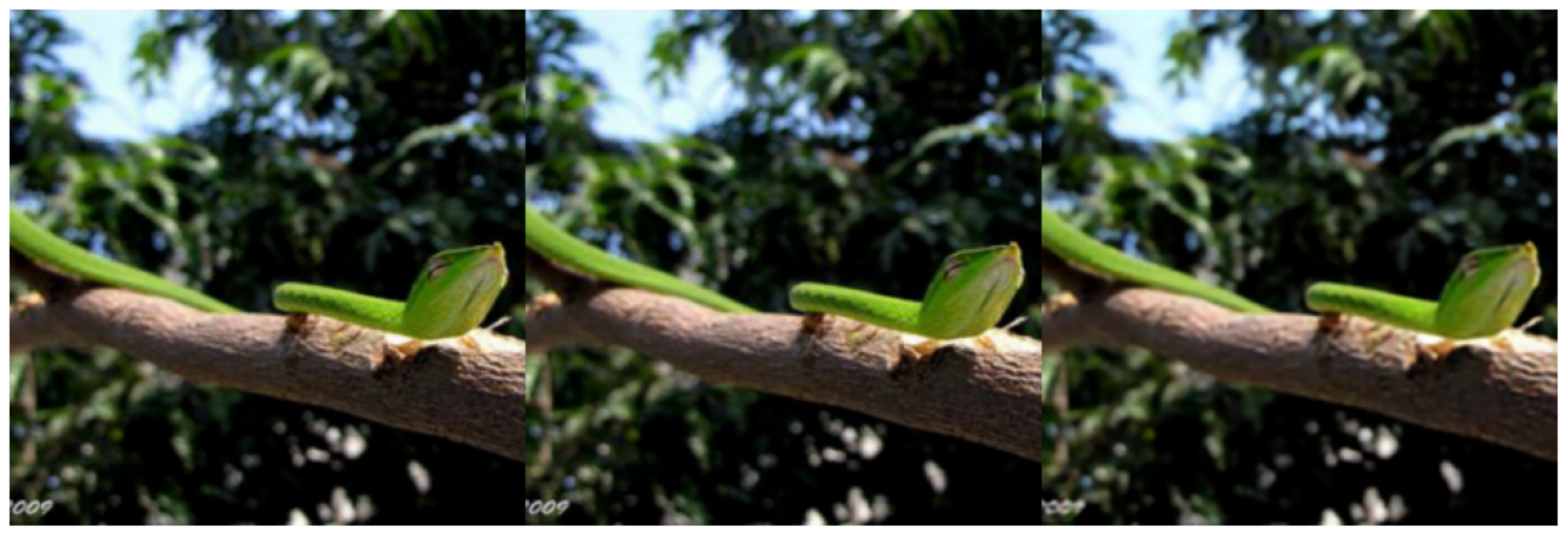}

     \end{subfigure}
     \begin{subfigure}[b]{.8\linewidth}
         \centering
         \includegraphics[width=.9\textwidth]{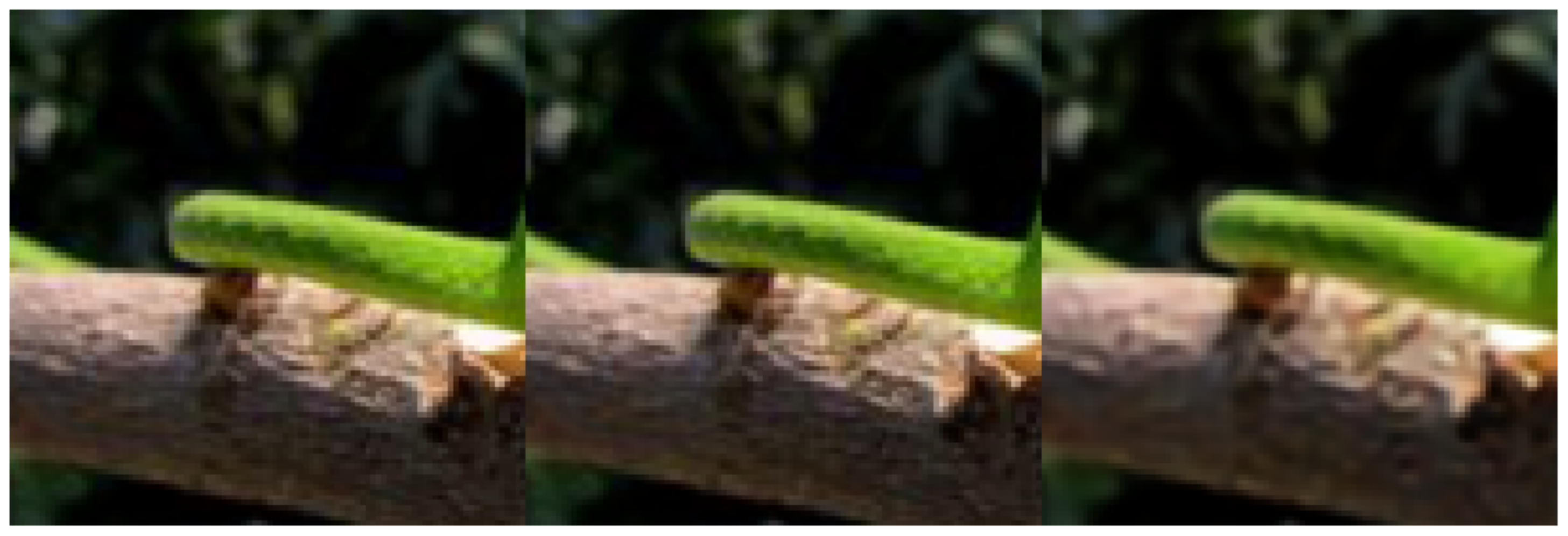}

     \end{subfigure}


     
         \begin{subfigure}[b]{.8\linewidth}
         \centering
         \includegraphics[width=.9\textwidth]{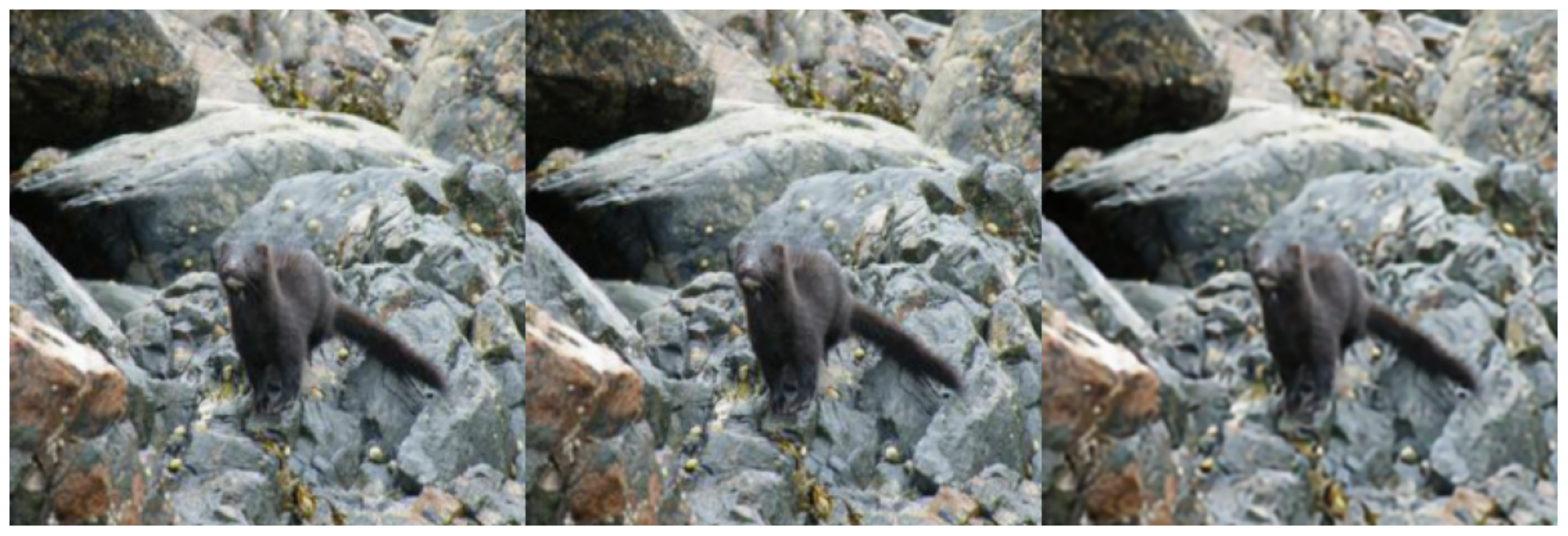}

     \end{subfigure}
         \begin{subfigure}[b]{.8\linewidth}
         \centering
         \includegraphics[width=.9\textwidth]{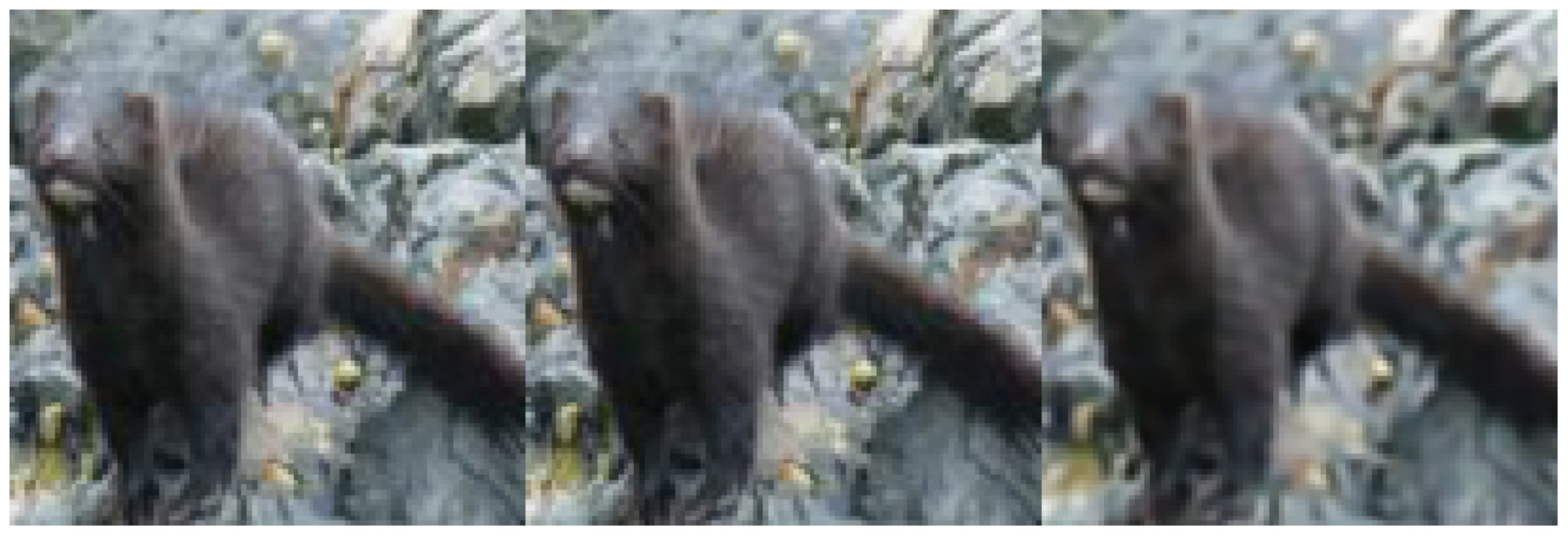}

     \end{subfigure}
                  \caption{The original image (left) is fed through and reconstructed by a model trained with DQ (middle) and VQ (right). The model is trained using identical settings. Perceptual quality of DQ outperforms VQ.}
\end{figure*}
\begin{figure*}[ht]
     \centering
         \begin{subfigure}[b]{\linewidth}
         \centering
         \includegraphics[width=1\textwidth]{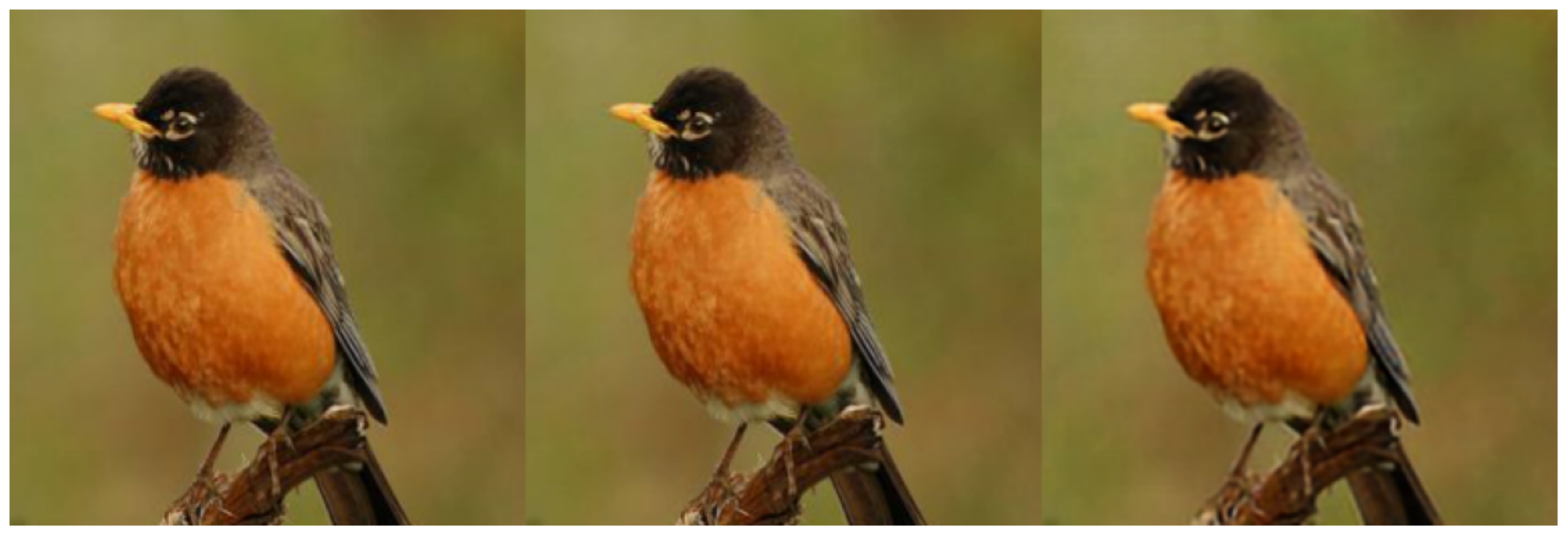}

     \end{subfigure}
         \begin{subfigure}[b]{\linewidth}
         \centering
         \includegraphics[width=1\textwidth]{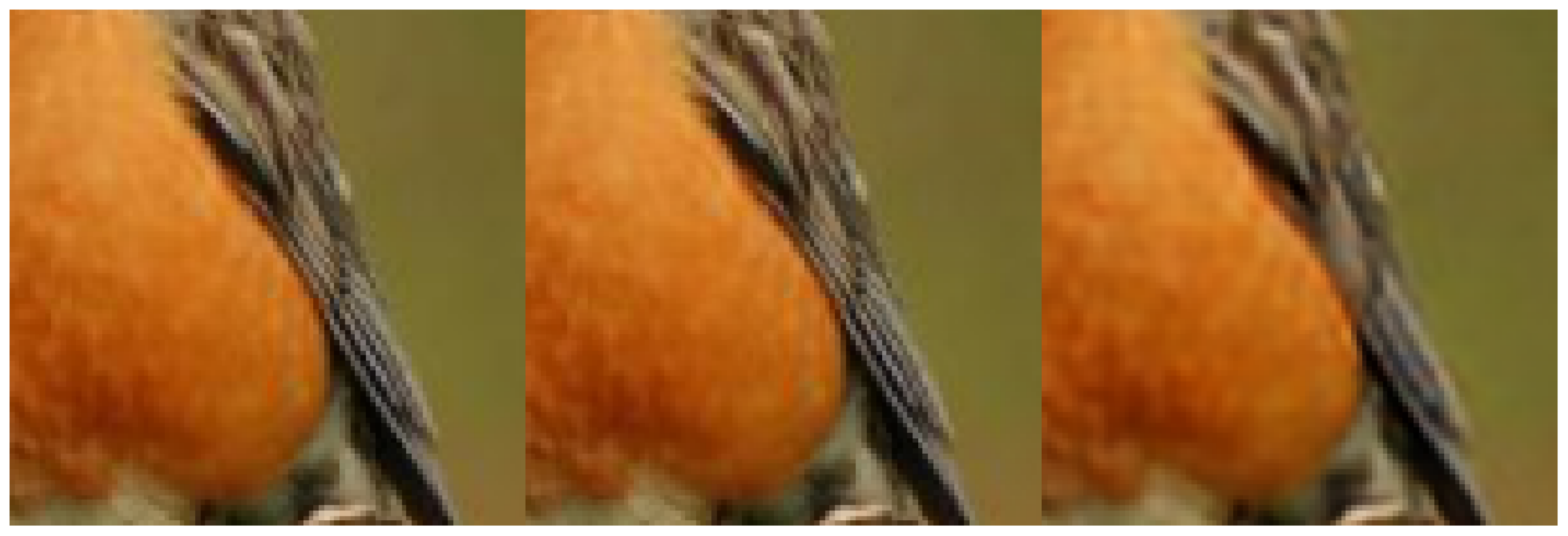}

     \end{subfigure}
         \begin{subfigure}[b]{\linewidth}
         \centering
         \includegraphics[width=1\textwidth]{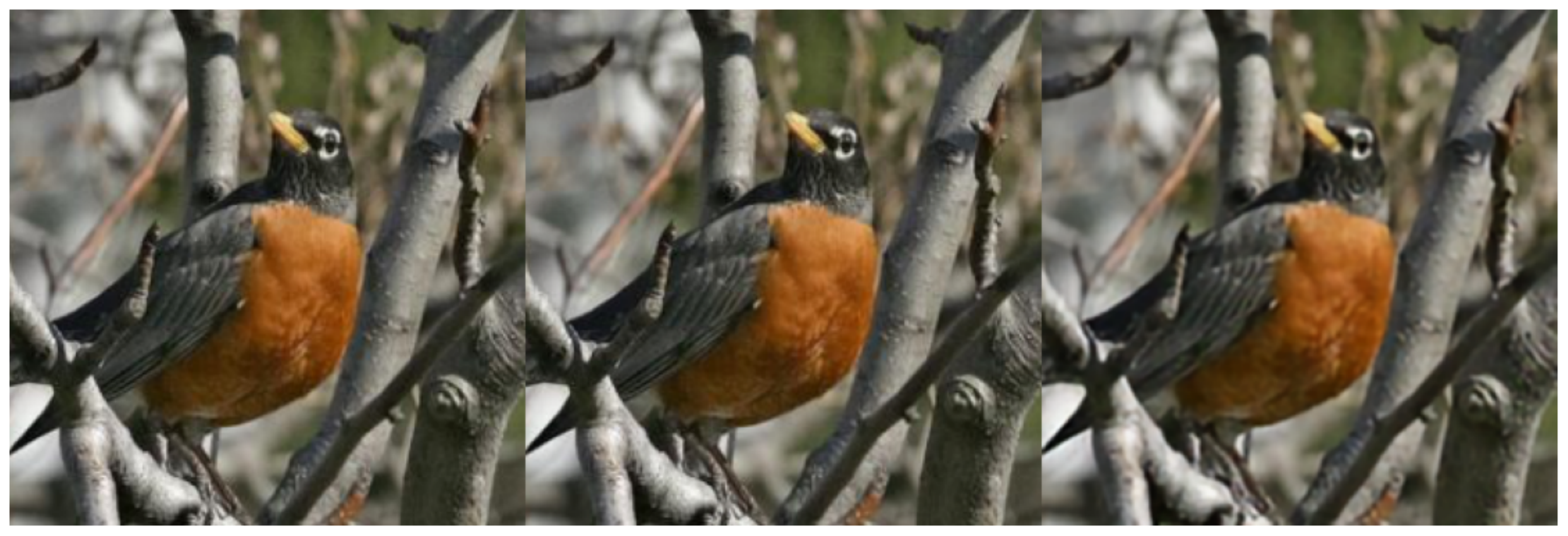}

     \end{subfigure}
         \begin{subfigure}[b]{\linewidth}
         \centering
         \includegraphics[width=1\textwidth]{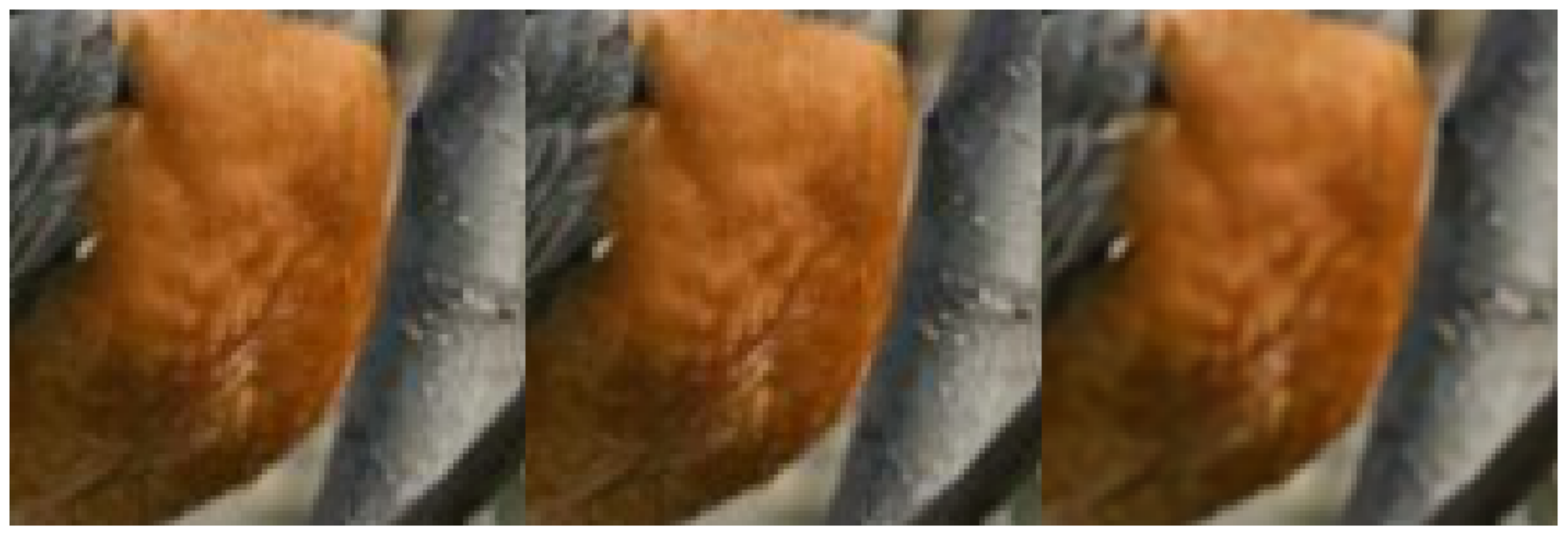}

     \end{subfigure}
         \caption{The original image (left) is fed through and reconstructed by a model trained with DQ (middle) and VQ (right). The model is trained using identical settings. Perceptual quality of DQ outperforms VQ.}
\end{figure*}

\end{document}